\RequirePackage{amsthm}
\documentclass[sn-mathphys-num]{sn-jnl}
\usepackage{graphicx}%
\usepackage{multirow}%
\usepackage{amsmath,amssymb,amsfonts}%
\usepackage[title]{appendix}%
\usepackage{xcolor}%
\usepackage{textcomp}%
\usepackage{manyfoot}%
\usepackage{booktabs}%
\usepackage{algorithmicx}%
\usepackage{algpseudocode}%
\usepackage{listings}%
\usepackage[section]{placeins}
\usepackage{float}
\usepackage{soul, color, xcolor}
\sethlcolor{yellow}
\usepackage{booktabs}
\usepackage{lmodern}
\usepackage{hyperref}
\usepackage{algorithm}%

\theoremstyle{thmstyleone}%
\theoremstyle{thmstyletwo}%

\theoremstyle{thmstylethree}%

\begin{document}

\title[Article Title]{Fast-Slow-Thinking: Complex Task Solving with Large Language Models}


\author[1,2,3]{\fnm{Yiliu} \sur{Sun}}\email{yiliusun@njust.edu.cn}

\author[1,2,3]{\fnm{Yanfang} \sur{Zhang}}\email{yanfangzhang@njust.edu.cn}

\author[1,2,3]{\fnm{Zicheng} \sur{Zhao}}\email{zicheng.zhao@njust.edu.cn}

\author[1,2,3]{\fnm{Sheng} \sur{Wan}}\email{wansheng315@hotmail.com}

\author[4]{\fnm{Dacheng} \sur{Tao}}\email{dacheng.tao@ntu.edu.sg}

\author*[1,2,3]{\fnm{Chen} \sur{Gong}}\email{chen.gong@njust.edu.cn}

\affil[1]{\orgdiv{School of Computer Science and Engineering}, \orgname{Nanjing University of Science and Technology}, \orgaddress{\city{Nanjing}, \postcode{210094}, \country{P.R. China}}}

\affil[2]{\orgdiv{Key Laboratory of Intelligent Perception and Systems for High-Dimensional Information of Ministry of Education}}

\affil[3]{\orgdiv{Jiangsu Key Laboratory of Image and Video Understanding for Social Security}}

\affil[4]{\orgdiv{The University of Sydney}, \orgaddress{\country{Australia}}}


\abstract{Nowadays, Large Language Models (LLMs) have been gradually employed to solve complex tasks. To face the challenge, task decomposition has become an effective way, which proposes to divide a complex task into multiple simpler subtasks and then solve them separately so that the difficulty of the original task can be reduced. However, the performance of existing task decomposition methods can be suboptimal when the task contains overly complex logic and constraints. In this situation, the solution generated by LLMs may deviate from the original purpose of the task, or contain redundant or even erroneous content. Therefore, inspired by the fact that humans possess two thinking systems including fast thinking and slow thinking, this paper introduces a new task decomposition method termed ``Fast-Slow-Thinking'' (FST), which stimulates LLMs to solve tasks through the cooperation of Fast Thinking (FT) and Slow Thinking (ST) steps. Here FT focuses more on the general and concise aspect of the task, and ST focuses more on the details of the task. In FT, LLMs are prompted to remove the constraints of the original task, therefore simplifying it to a general and concise one. In ST, we recall the constraints removed in FT, so that LLMs can improve the answer generated in FT to meet the requirements of the original task. Therefore, our FST method enables LLMs to consider a complex problem via a human-like cognition process from coarse to fine, the effectiveness of which has been well demonstrated by the experiments on three types of tasks.}




\maketitle

\section{Introduction}
\label{introduction}

The pre-trained Large Language Models (LLMs) have attracted widespread attention in many fields related to natural language processing, such as GPT \cite{achiam2023gpt, brown2020language, radford2018improving}, LLaMA \cite{touvron2023llama}, and Gemini \cite{team2023gemini}. So far, numerous efforts have been made to enhance the performance of LLMs, including fine-tuning \cite{dettmers2024qlora, houlsby2019parameter, ludan2023explanation, xu2023multiinstruct}, prompt engineering \cite{wang2023towards, wang2022self, wang2023element}, and mixture-of-experts (MoE) \cite{du2022glam, fedus2022switch, gupta2024offline, jacobs1991adaptive, jain2024damex}. While these techniques have enabled LLMs to perform well on some simple tasks, their effectiveness is still inadequate for complex tasks. To address this issue, task decomposition has become an effective strategy to improve the performance of LLMs on complex tasks, which breaks down a complicated task into simple subtasks and then solves them separately. Existing task decomposition methods \cite{besta2024graph, chen2024boosting, khot2021text, khot2023decomposed, lei2023boosting, patel2022question, yao2024tree} have demonstrated promising results on the tasks such as arithmetic reasoning and commonsense reasoning.

\begin{figure*}[t]
  \includegraphics[width=\linewidth]{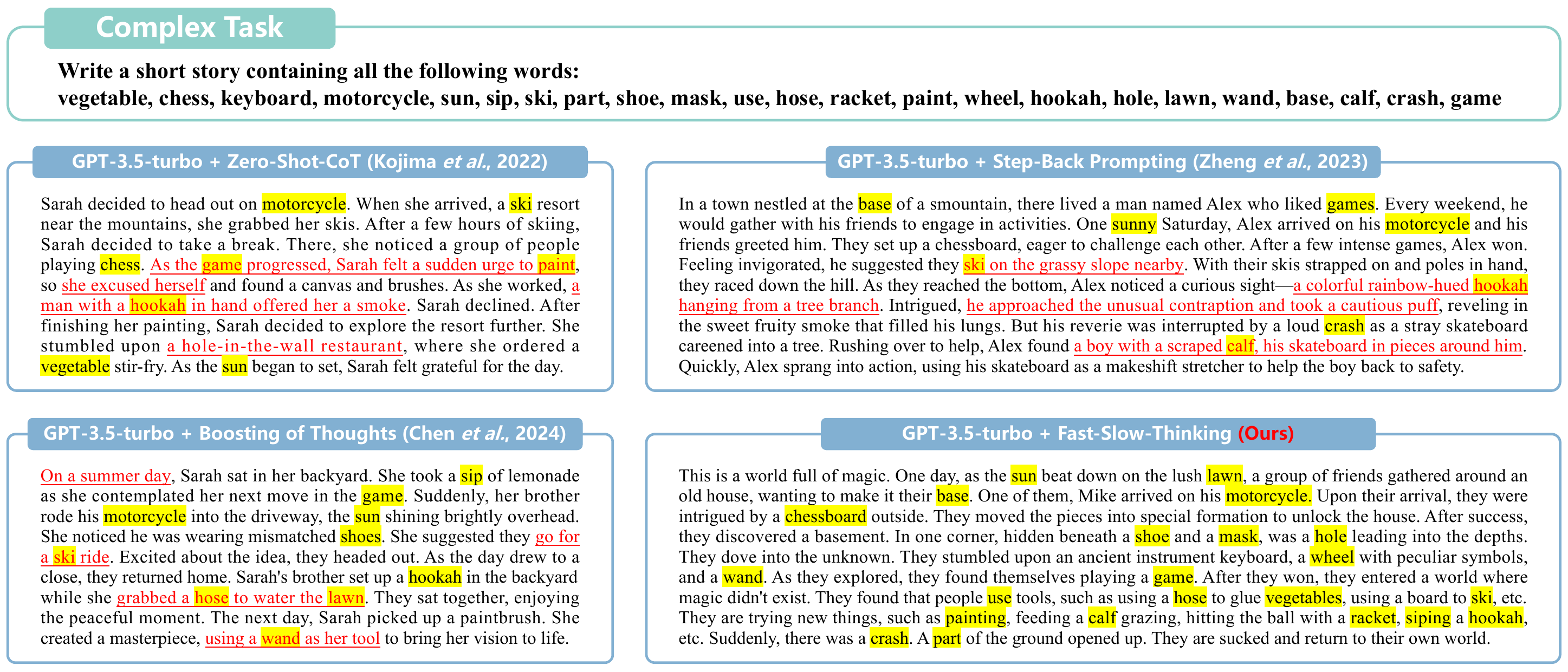}
  \caption {Comparison between our method and other typical task decomposition methods. The texts highlighted in yellow denote the words that must be included in the story, and the red and underlined texts denote the illogical content. The story generated by FST contains all the specified words, and the content is also reasonable.}
  \label{figure1}
\end{figure*}

Unfortunately, if the task contains excessively complex constraints, the existing task decomposition methods can result in unreasonable decomposition. This could lead to solutions that deviate from the original purpose of the task or contain redundant and irrelevant content. Taking the task shown in Figure~\ref{figure1} as an example, the stories generated by existing task decomposition methods contain a limited number of specified words and include illogical content. It is because LLM pays too much attention to the constraints of story generation, and thus ignores the commonsense and readability that the story needs to have.

To enable LLMs to deal with complex tasks, this paper proposes a new task decomposition method termed ``Fast-Slow-Thinking'' (FST). This is inspired by the human thinking mode revealed by \cite{kahneman2017thinking}, which divides the human thinking mode into fast thinking and slow thinking systems. As described in \cite{kahneman2017thinking}, the fast thinking system aims to quickly grasp an overall perception of the task, and the slow thinking system aims to carefully analyze details to generate a solution meeting the requirements of the task. Unlike existing task decomposition methods that encourage LLMs to primarily focus on the details of the task, which are similar to the slow thinking system, FST mimics the human thinking mode by combining the fast thinking and slow thinking systems. Specifically, in the first step termed ``Fast Thinking'' (FT), appropriate prompts are designed for LLMs to remove the complex and fussy constraints from the task, so as to simplify the task to a concise and general one. In the second step termed ``Slow Thinking'' (ST), LLMs are prompted to reconsider the constraints of the original task and improve the solution generated in FT to meet the constraints. To reduce the potential mistakes made in FT and ST, we introduce the third step termed ``Output Inspection'' (OI), in which prompts are designed to stimulate LLMs to carefully review and correct the improper answer. Through these three steps, LLMs are expected to accurately and quickly grasp the main idea of a complex task without ignoring any details. As demonstrated in Figure~\ref{figure1}, the story generated by ``GPT-3.5-turbo + Fast-Slow-Thinking'' contains all specified words, maintains a complete storyline, and exhibits reasonable logic.

The effectiveness of our FST method has been well confirmed by the experiments on diverse tasks including math reasoning, long-content answering, and constrained story generation, where GPT-3.5-turbo, Llama-3.1-8B-Instruct, and Gemini-pro are adopted as backbone LLMs. For instance, in the math reasoning task, FST increases \textit{Result Accuracy} (RA) by up to 15.85\% on GSM8K dataset and 13.85\% on MMLU-pro datasets, respectively. In the long-content answering task, FST improves the understanding ability of the models and enables them to generate more accurate answers than other baseline methods, which achieves a maximal \textit{Macro-average} (M-Avg) of 52.59\% on LongBench dataset and a maximal \textit{Quadratic Weighted Kappa} (QWK) score of 0.679 on ASAP dataset. In the constrained story generation task, FST increases \textit{All Present Rate} (APR) of the adopted backbone LLMs by up to 10\% on CommonGen-Hard dataset.

It is worth clarifying that although some existing works \cite{yao2024tree, zhang2023cumulative} also mention the two systems of human thinking mode (\textit{i.e.}, fast thinking and slow thinking), they only refer to these two systems implicitly rather than directly incorporate them into algorithm design. Differently, our FST is the first work that explicitly combines the two systems for problem decomposition, and thereby improves the performance of LLMs. Besides, while FST appears similar to Step-Back Prompting \cite{zheng2024take} since both methods adopt a general-to-specific order, they differ markedly in their implementations. To be concrete, Step-Back Prompting focuses on the abstraction or summarization of common rules and principles necessary to complete the task. Differently, FST operates in a coarse-to-fine manner, where the general task extracted in FT is the coarsening or simplification of the original task. This task is further refined by adding detailed constraints or requirements. For example, for the task ``build a three-story villa in the countryside'', Step-Back Prompting would pay attention to ``what are the skills and knowledge involved in completing this task'' Unlike Step-Back Prompting, the proposed FST will first simplify it to ``build a villa'', and then consider the excessively complex constraints such as ``the villa is three-story'' and ``the villa is built in the countryside''

\section{Related Work}
\label{related}

This section investigates task decomposition, which is fundamental and critical in improving the performance of LLMs on complex tasks. According to \citep{saha2024branch}, the existing task decomposition methods can be categorized as \textit{sequential decomposition methods} and \textit{parallel decomposition methods}, which will be reviewed as follows.

\textbf{Sequential Decomposition Methods.} These methods \citep{chen2024boosting, dua2022successive, khot2021text, khot2023decomposed, patel2022question, wang2023plan, wei2022chain, yao2024tree, zhouleast} aim to solve reasoning problems, where the solutions contain sequential relationships, such as mathematical reasoning and commonsense reasoning. For instance, Kojima \textit{et al.} \citep{kojima2022large} design a simple but effective prompt ``Let’s think step by step'' to form a zero-shot Chain-of-Thought (Zero-Shot-CoT), which stimulates LLMs to solve the problems step by step. Based on Zero-Shot-CoT, Plan-and-Solve (PS) Prompting \citep{wang2023plan} adds more details to the prompt template to reduce errors in the intermediate reasoning processes. While these works can improve the performance of LLMs on various tasks, they face a dilemma that the reliability of the subtasks obtained through decomposition remains uncertain. Therefore, many works \citep{chen2024boosting, dua2022successive, khot2021text, yao2024tree, zhang2023cumulative} aim to explore the ways to iteratively decompose the tasks. For example, Yao \textit{et al.} \citep{yao2024tree} proposes Tree of Thoughts (ToT), which explores different ways to solve the task and chooses the optimal one in each step. Though similar to ToT, the Cumulative Reasoning method \citep{zhang2023cumulative} is distinguished by its ability to store and exploit the historically correct subtasks, which can guide subsequent decomposition steps. Moreover, Chen \textit{et al.} \citep{chen2024boosting} proposes Boosting of Thoughts (BoT), which also stores historically incorrect subtasks to assist LLMs in generating correct subsequent subtasks.

\textbf{Parallel Decomposition Methods.} A few works \citep{besta2024graph, lei2023boosting, saha2024branch} focus on the tasks where subtasks are independent of each other, such as essay evaluation and constrained story generation. Among them, Graph-of-Thoughts (GoT) \citep{besta2024graph, lei2023boosting} models the reasoning processes of LLMs as an arbitrary graph, where the steps to solve the tasks are vertices and edges correspond to the relationships between these vertices. The difference between \citep{besta2024graph} and \citep{lei2023boosting} is that the former considers tasks like number sorting and keyword counting, while the latter considers tasks like 24-point game and resolution of high-order equations. Based on the idea of GoT, Saha \textit{et al.} \citep{saha2024branch} introduces the Branch-Solve-Merge (BSM) method, which is a fixed program designed to solve complex tasks like constrained story generation. As the execution trail of BSM forms a graph, BSM can be seen as a special case of GoT.

As mentioned in Section~\ref{introduction}, most of these works encourage LLMs to explore the details of a given complex task, which exactly corresponds to the slow thinking system of the human thinking mode. Therefore, task decomposition can sometimes be difficult as LLMs may pay too much attention to the details and ignore the global meaning of the task. Moreover, there are a few works \citep{christakopoulou2024agents, lin2024swiftsage, saha2024system, pan2024dynathink} involving fast and slow thinking systems in the task solving process. However, they view the fast and slow thinking systems as two separate systems when facing a task. When the fast thinking system is unable to solve the task, it is handed over to the slow thinking system for resolution. Therefore, their implementations tend to train two separate models, where the model with fewer parameters represents the fast thinking system and the model with more parameters represents the slow thinking system. Differently, our proposed FST views the two thinking systems as a progressive relationship. When facing a task, FST focuses on different aspects of the task in different thinking systems. The content generated in the fast thinking system will help the slow thinking system generate a better solution to the task. In other words, it integrates the merits of fast thinking and slow thinking systems in a coarse-to-fine manner. As a consequence, both the global and local information of a task can be considered, which yields better performance than previous methods.

\begin{figure*}[t]
  \includegraphics[width=\linewidth]{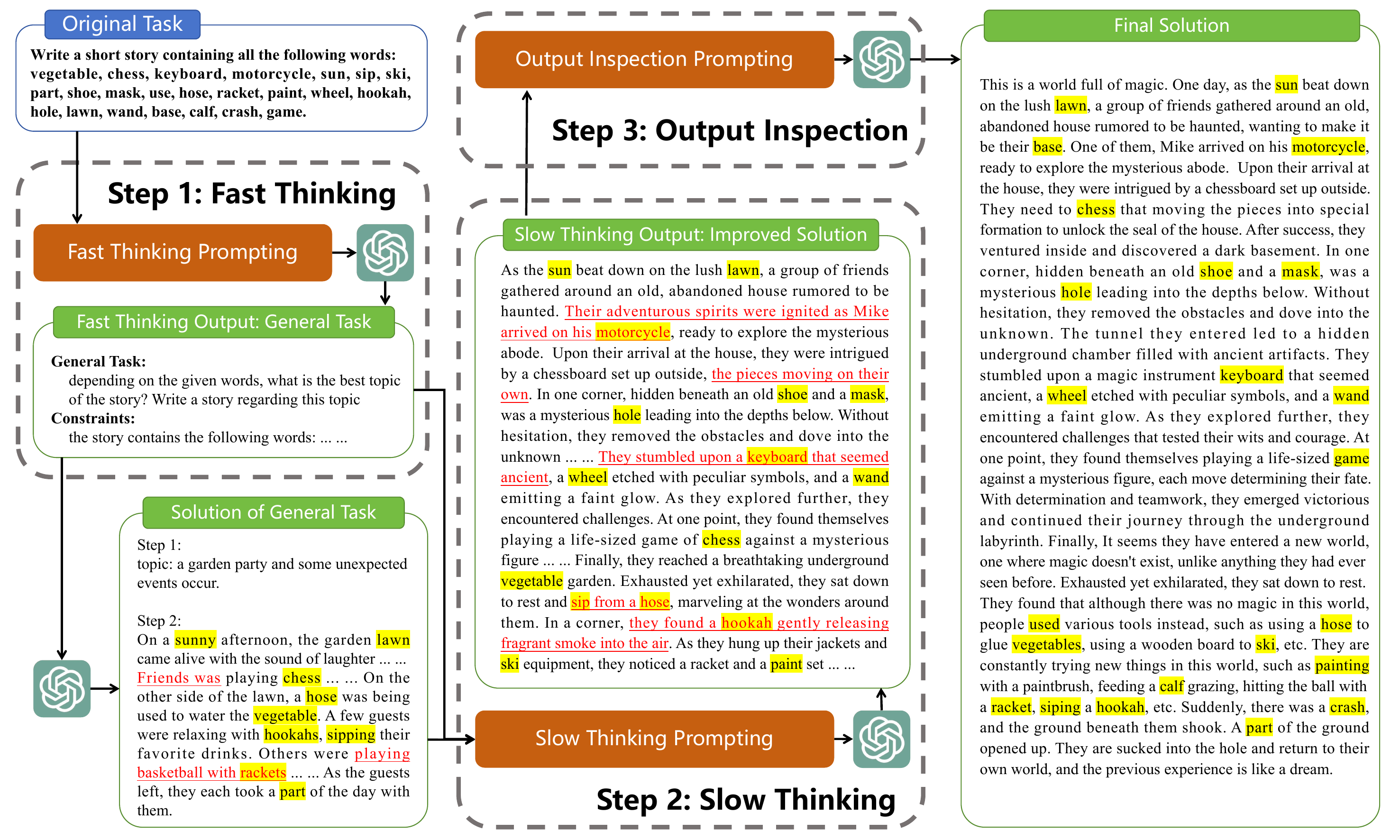}
  \caption {An illustration of our Fast-Slow-Thinking, which is inspired by the human thinking mode and consists of three steps: Fast Thinking, Slow Thinking, and Output Inspection. The texts highlighted in yellow denote the words that must be included in the story, and the red and underlined texts denote the illogical content. We prompt LLMs to simplify the original task to a concise and general one and complete the concise and general task in FT. In ST, LLMs need to reconsider the original task and improve the solution generated in FT to meet the requirements. In OI, LLMs are prompted to check the improved solution for potential mistakes.}
  \label{figure2}
\end{figure*}

\section{Methodology}
\label{methodology}

In this section, we introduce the principle of our proposed Fast-Slow-Thinking (FST) in detail, which comprises three steps: Fast Thinking (Section~\ref{fast}), Slow Thinking (Section~\ref{slow}), and Output Inspection (Section~\ref{ouput}). Unlike the existing task decomposition methods that resemble the slow thinking system, FST integrates both fast and slow thinking to enable LLMs to quickly and accurately grasp the main idea of complex tasks without overlooking any details.

\subsection{Fast Thinking}
\label{fast}

For humans, the fast thinking system serves as the initial pre-processing system when confronting complex tasks. In this system, humans tend to simplify the original task while retaining its main idea, enabling a quick overall understanding. As shown in Figure~\ref{figure2} (see ``Step 1: Fast Thinking''), we first prompt LLMs to remove complex constraints of the task, so that the complex task can be simplified to a concise and general one. Then, LLMs are prompted to generate a solution for this simplified task, which can then serve as a guideline and framework for tackling the original complex task. For example, if the original task is ``write a story containing words: \textless words\textgreater'', the concise and general task involves generating a story belonging to a topic that is relevant to the specified words. If the original task is ``\textless essay\textgreater \ evaluate the essay following the protocols: \textless protocols\textgreater'', the concise and general task is freely evaluating the essay without strictly following the specified protocols.

Moreover, FT has another potential benefit. Different from most of the task decomposition methods which encourage LLMs to deal with the entire complex task directly, the FT step operates in a fine-to-coarse manner. As a result, the complex task can be transformed into a form that is similar to the previously observed training data of the models. For example, LLMs may have never seen problems like ``build a three-story villa in the countryside, with a budget of less than \$100,000'', but LLMs may have been trained with the data like ``build a villa'' In this case, simplification to a concise and general task can better stimulate the power of LLMs.

\subsection{Slow Thinking}
\label{slow}

Although the solution generated by the fast thinking system provides a broad overview of the task, it is not always optimal as it often ignores important details. To address this, we deploy a slow thinking step which aims to solve the original task thoroughly and refine the preliminary solution created during the fast thinking step. Consequently, as illustrated in ``Step 2: Slow Thinking'' of Figure~\ref{figure2}, we have designed the second step named ST to prompt LLMs to take a deeper thinking. In ST, under the guidance of the overall understanding from FT, LLMs are prompted to focus on the constraints or requirements that are disregarded in FT and refine the solution accordingly to make it comprehensive and precisely aligned with the requirements of the original task. 

For different tasks, the constraints considered in ST vary depending on the operation in FT. We revisit the examples mentioned in Section~\ref{fast}. For the task ``write a story containing words: \textless words\textgreater'', LLMs need to consider how to refine the story generated in FT to ensure that it contains all specified words. For the task ``\textless essay\textgreater \ evaluate the essay following the protocols: \textless protocols\textgreater'', LLMs need to comprehensively consider the preliminary evaluation results generated in FT and score the essay in strict accordance with the specified protocols.

\subsection{Output Inspection}
\label{ouput}

To avoid the potential mistakes that appear in the FT and ST steps, we further introduce an Output Inspection (OI) step (see ``Step 3: Output Inspection'' in Figure~\ref{figure2}). To verify the correctness of the generated content, such OI step primarily focuses on two aspects: Firstly, whether the solution strictly fulfills the requirements of the task. If not, LLMs need to improve the solution; Secondly, whether the solution fulfills some other requirements, such as intermediate process and content format. If not, LLMs need to further refine the solution.

\section{Algorithm Implementation}
\label{implementation}

The above section presents the general principles of our proposed FST algorithm. In this section, we introduce the identical prompt template of the proposed FST that is applicable to different tasks. For different complex tasks, the specific prompts of FST follow the prompt template mentioned above. The overview of prompt template are shown in Table~\ref{prompt_template_FST}.

It is worth mentioning that the design of the prompt template strictly follows the principle introduced in Section~\ref{methodology}. When applied to different tasks, only some simple content needs to be replaced, such as the content of the task and the domain of the task. Therefore, our proposed FST is a general method that can be easily applied to different types of tasks.

\subsection{Fast Thinking}

As introduced in Section~\ref{fast}, the Fast Thinking step tends to simplify the original complex task while retaining its main idea. Therefore, the implementation of the Fast Thinking step can be designed into three parts, namely \textit{Identity Setup}, \textit{Task Simplification}, and \textit{Answer Generation}.

\textbf{Identity Setup.} This part aims to set the mindset of LLMs to that of the expert in the specific domain. In this way, LLMs can better understand the tasks of this domain. The prompt template of the Identity Setup part is ``You are an expert in \textless the domain of the task\textgreater. You are very good at understanding and solving the tasks in this domain.''

\textbf{Task Simplification.} This part aims to simplify the complex task into a concise and general one. We use few-shot prompts to reach the target. The prompt template is ``Now I will give you a complex task, \textless the task\textgreater. You need to understand the task and simplify the task into a concise and general one. I give you some simplification examples below as guidance: \textless simplification examples\textgreater.''

\textbf{Answer Generation.} This part aims to generate the answer to the concise and general task. Therefore, the prompt template is ``Then, please generate the answer to the concise and general task.''

\subsection{Slow Thinking}

As introduced in Section~\ref{slow}, the Slow Thinking step tends to reconsider the constraints removed in the Fast Thinking step and improve the answer to the concise and general task. Therefore, the implementation of the Slow Thinking step can be designed into two parts, namely \textit{Constraint Reconsideration} and \textit{Answer Improvement}.

\textbf{Constraint Reconsideration.} This part aims to stimulate LLMs to reconsider the constraints of the original task. In this way, LLMs take a deep understanding of the complex task without deviating from the main idea of the task. The prompt template of the Constraint Reconsideration part is ``Based on the concise and general task \textless the concise and general task\textgreater, I will add some constraints: \textless the constraints removed in the Fast Thinking step\textgreater. Please take a deep consideration of these constraints.''

\textbf{Answer Improvement.} This part aims to improve the answer to the concise and general task by considering the constraints carefully. Therefore, the prompt template for the Answer Improvement part is ``Please improve the answer of the concise and general task to meet these constraints.'' Moreover, some general tips can be added to provide specific guidance, like ``You should pay attention to the intermediate process.'', ``The logic must be reasonable'', and so on.

\begin{table*}[t]
\caption{\label{prompt_template_FST}
    Specific Prompt Template of FST.
  }
  \centering
  \small
\resizebox{\columnwidth}{!}{
\begin{tabular}{cl}
\toprule
Step                   & \multicolumn{1}{c}{Prompt Template}                                           
\\ \midrule
FT                         & \begin{tabular}[c]{@{}p{\textwidth}@{}} You are an expert in \textless{}the domain of the task\textgreater{}. You are very good at understanding and solving the tasks in this domain. Now I will give you a complex task, \textless{}the task\textgreater{}. Your work should follow the two steps below: \\ \\Step 1: You need to understand the task and simplify the task into a concise and general one. \\I give you some simplification examples below as guidance: \\        \textless{}simplification examples\textgreater\\ \\ Step 2: Please generate the answer to the concise and general task. \end{tabular}                                                                                                                                                                                                                                                                                                   \\  \midrule
ST                         & \begin{tabular}[c]{@{}p{\textwidth}@{}} Based on the concise and general task \textless{}the concise and general task\textgreater{}, I will add some constraints:\\ \textless{}the constraints removed in the Fast Thinking step\textgreater{}\\ Please take a deep consideration of these constraints and improve the answer of the concise and general task below to meet these constraints. \\ \textless the answer of the concise and general task\textgreater \\ \\ Tips: \\ 1. Pay attention to the correctness of the intermediate process.\\ 2. The logic must be reasonable.
\end{tabular} \\  \midrule

OI & \begin{tabular}[c]{@{}p{\textwidth}@{}} \textless{}the original task\textgreater\\ The answer is: \textless{}the answer generated in the Slow Thinking step\textgreater\\ \\ You need to check the answer through the following steps:\\ Step 1: Whether the answer strictly meets the requirements of the task. If not, please improve it.\\ Step 2: Whether the intermediate process is correct. If not, please improve it.\\ Step 3: Can every sentence of the answer be supported by the problem and material given to you? If not, modify unsupported parts. \end{tabular}                                                           \\ \bottomrule
\end{tabular}}
\end{table*}

\subsection{Output Inspection}

As introduced in Section~\ref{ouput}, the Output Inspection step tends to verify the correctness of the answer generated in the step of Slow Thinking. Moreover, some complex tasks ask for some other characteristics in the answers, such as smooth presentation, detailed intermediate process, and so on. Therefore, the implementation of the Output Inspection step can be designed into two parts, namely \textit{Correctness Check} and \textit{Other Check}.

\textbf{Correctness Check.} This part aims to check whether the answer generated in the Slow Thinking step meets the requirements of the original complex task. The prompt template of the Correctness Check part is ``You need to check whether the answer strictly meets the requirements of the task. If not, please improve the answer.''

\textbf{Other Check.} This part aims to stimulate LLMs to consider some other characteristics, which are needed to be contained in the answer. In our implementation, We consider some general points applicable to different types of tasks. To further improve the task performance of FST, some task-specific checks can be added in this part. For example, for the math tasks, the answers need to pay attention to the calculation of the intermediate process. For the constrained text generation tasks, the presentations need to be smooth. However, these task-specific checks are not considered in our experiments.

\begin{table*}[t]
  \centering
  \small
  \caption{\label{experimental_data}
    Statistics of the datasets used in the experiments.
  }
  \setlength{\tabcolsep}{4mm}{
\begin{tabular}{ccc}
\toprule
Task                                    & Dataset        & Number of Examples \\ \midrule
\multirow{2}{*}{Math Reasoning}         & GSM8K \citep{cobbe2021training}          & 2000               \\
                                        & MMLU-pro \citep{wang2024mmlu}           & 1351 (All)          \\ \midrule
\multirow{2}{*}{Long-Content Answering} & LongBench \citep{bai2023longbench}      & 2550               \\
                                        & ASAP \citep{ASAP}           & 800                \\ \midrule
Constrained Story Generation            & CommonGen-Hard \citep{madaan2024self} & 200 (All)          \\ \bottomrule
\end{tabular}}
  
\end{table*}

\section{Experiments}
\label{experiments}

This part introduces the tasks, backbone models, baseline methods, and evaluation metrics used in our experiments. Moreover, we present the experimental results and detailed analysis to highlight the effectiveness of our proposed FST.

\textbf{Tasks.} In our experiments, we consider three typical complex tasks, namely 1) math reasoning, 2) long-content answering, and 3) constrained story generation. The datasets used in three tasks are GSM8K, MMLU-pro, LongBench, ASAP, and CommonGen-Hard datasets, which are introduced in detail in Sections~\ref{mathreasoning}, \ref{longcontentanswering}, and \ref{constrainedstorygeneration}. It is worth mentioning that the data used for the experiments of three tasks is randomly sampled from the original datasets. Table~\ref{experimental_data} illustrates the number of examples and subtasks utilized for the experiments on the five datasets mentioned above.

\begin{table}[t]
\renewcommand{\arraystretch}{1.1}
\centering
  \caption{\label{experiment_result}
    Comparison results on the three different types of tasks. Here, ``$\uparrow$'' means that higher values are better, ``$\downarrow$'' means that lower values are better, and ``--'' means that the method is not applicable to this type of task. The best records under each metric are highlighted in bold. 
  }
 \setlength{\tabcolsep}{1mm}{
\begin{tabular}{lccccccc}
\hline
\multicolumn{1}{c}{\multirow{2}{*}{Method}} & GSM8K            & MMLU-pro         & LongBench        & ASAP           & \multicolumn{3}{c}{CommonGen-Hard}                     \\ \cmidrule{2-8}
\multicolumn{1}{c}{}                        & RA $\uparrow$               & RA $\uparrow$               & M-Avg $\uparrow$            & QWK $\uparrow$            & APR $\uparrow$              & MCR $\downarrow$              & OSR $\uparrow$              \\ \hline
GPT-3.5-turbo                               & 73.30\%          & 39.53\%          & 44.00\%          & 0.232          & 2.00\%           & 18.00\%          & 0.50\%           \\
\ \ + Zero-Shot-CoT \citep{kojima2022large}                      & 81.65\%          & 41.75\%          & 46.45\%          & 0.286          & 3.00\%           & 17.80\%          & 2.00\%           \\
\ \ + BoT \citep{chen2024boosting}                                      & 87.35\%          & 43.60\%          & 49.67\%          & 0.365          & 6.00\%           & 21.08\%          & 11.00\%          \\
\ \ + SPP \citep{wang2024unleashing}                                      & 84.45\%          & 43.15\%          & 47.37\%          & 0.379          & 4.50\%           & 15.30\%          & 15.00\%          \\
\ \ + Step-Back \citep{zheng2024take}                                & 81.35\%          & 42.19\%          & 47.04\%          & 0.335          & 3.00\%           & 17.30\%          & 3.50\%           \\
\ \ + DynaThink \citep{pan2024dynathink}                                & 85.85\%          & \textbf{49.96\%}          & --          & 0.597          & --           & --          & --           \\
\ \ + PS \citep{wang2023plan}                                & 83.65\%          & 42.26\%          & --          & --          & --           & --          & --           \\
\ \ + BSM \citep{saha2024branch}                                & --          & --          & --          & --          & 8.00\%           & 14.68\%          & 20.50\%           \\
\ \ + FST (\textbf{ours})                                & \textbf{88.65\%} & 45.97\% & \textbf{52.59\%} & \textbf{0.679} & \textbf{12.00\%} & \textbf{12.59\%} & \textbf{47.50\%} \\ \hline
Llama-3.1-8B-Instruct                       & 72.50\%          & 22.35\%          & 35.64\%          & 0.181          & 1.00\%           & 22.39\%          & 3.50\%           \\
\ \ + Zero-Shot-CoT \citep{kojima2022large}                             & 84.50\%          & 29.02\%          & 37.95\%          & 0.206          & 2.00\%           & 20.45\%          & 5.00\%           \\
\ \ + BoT \citep{chen2024boosting}                                      & 86.25\%          & 34.79\%          & 42.04\%          & 0.305          & 4.00\%           & 18.52\%          & 10.50\%          \\
\ \ + SPP \citep{wang2024unleashing}                                      & 85.95\%          & 32.49\%          & 40.49\%          & 0.279          & 2.50\%           & 19.34\%          & 10.00\%          \\
\ \ + Step-Back \citep{zheng2024take}                                & 85.85\%          & 31.09\%          & 40.21\%          & 0.231          & 2.50\%           & 20.01\%          & 5.50\%           \\
\ \ + DynaThink \citep{pan2024dynathink}                                & 85.90\%          & \textbf{37.01\%}          & --          & 0.513          & --           & --          & --           \\
\ \ + PS \citep{wang2023plan}                                & 85.75\%          & 33.60\%          & --          & --          & --           & --          & --           \\
\ \ + BSM \citep{saha2024branch}                                & --          & --          & --          & --          & 5.00\%           & 18.06\%          & 18.50\%           \\
\ \ + FST (\textbf{ours})                                & \textbf{88.05\%} & 36.20\% & \textbf{46.39\%} & \textbf{0.583} & \textbf{7.50\%}  & \textbf{16.14\%} & \textbf{48.00\%} \\ \hline
Gemini-pro                                  & 69.65\%          & 28.87\%          & 40.43\%          & 0.218          & 1.00\%           & 20.17\%          & 2.00\%           \\
\ \ + Zero-Shot-CoT \citep{kojima2022large}                            & 75.00\%          & 33.38\%          & 42.38\%          & 0.249          & 3.00\%           & 18.70\%          & 2.50\%           \\
\ \ + BoT \citep{chen2024boosting}                                      & 83.35\%          & 36.05\%          & 46.64\%          & 0.391          & 5.00\%           & 22.58\%          & 10.00\%          \\
\ \ + SPP \citep{wang2024unleashing}                                      & 79.65\%          & 35.16\%          & 45.19\%          & 0.361          & 4.00\%           & 17.26\%          & 13.00\%          \\
\ \ + Step-Back \citep{zheng2024take}                                & 77.15\%          & 33.90\%          & 44.02\%          & 0.279          & 3.00\%           & 19.12\%          & 6.00\%           \\
\ \ + DynaThink \citep{pan2024dynathink}                                & 81.35\%          & 39.45\%          & --          & 0.536          & --           & --          & --           \\
\ \ + PS \citep{wang2023plan}                                & 78.05\%          & 33.75\%          & --          & --          & --           & --          & --           \\
\ \ + BSM \citep{saha2024branch}                                & --          & --          & --          & --          & 8.00\%           & 16.58\%          & 22.50\%           \\
\ \ + FST (\textbf{ours})                                & \textbf{85.50\%} & \textbf{40.04\%} & \textbf{49.11\%} & \textbf{0.601} & \textbf{9.00\%}  & \textbf{14.52\%} & \textbf{44.00\%} \\ \hline
\end{tabular}}
\end{table}

\textbf{Baseline Methods.} Our proposed FST is compared with five different methods: 1) Backbone Models, 2) zero-shot Chain-of-Thought (Zero-Shot-CoT) \citep{kojima2022large}, 3) Boosting of Thoughts (BoT) \citep{chen2024boosting}, 4) Solo Performance Prompting (SPP) \citep{wang2024unleashing}, and 5) Step-Back Prompting \citep{zheng2024take}. Among them, Zero-Shot-CoT is a representative work of prompt design, which adds ``Let’s think step by step'' after the task to prompt LLMs to solve the problem step by step. BoT and SPP are two implementations that mimic the slow thinking system of human thinking mode. Differently, while BoT stimulates LLMs to iteratively explore and self-evaluate different reasoning processes in order to acquire an ensemble of trial-and-error reasoning experiences, SPP asks LLMs to play multiple roles, which will have multiple rounds of discussions about the task when facing a task. We choose Step-Back Prompting for comparison to demonstrate the superiority of our proposed FST, which inspires LLMs to find the knowledge and skills needed to solve the problem in the first step, and then solve the problem grounded on these knowledge and skills. Besides, our FST is compared with DynaThink~\citep{pan2024dynathink}, which combines fast and slow thinking systems in the reasoning process. It categorizes tasks into two distinct pathways (``Fast'' and ``Slow'') according to the multiple performances of LLMs on them. Since the algorithm design of DynaThink requires voting on multiple answers generated by LLMs, it is not applicable to LongBench and CommonGen-Hard datasets, which do not have fixed answers, but rather impose some constraints and semantic requirements on the content. Moreover, since existing works such as Plan-and-Solve (PS) \citep{wang2023plan} and Branch-Solve-Merge (BSM) \citep{saha2024branch} are specifically designed for math reasoning task and constrained story generation task, respectively, here we also compare our FST with these two task-specific methods on the corresponding tasks. The implementation details of BoT, SPP, Step-Back Prompting, PS, BSM, and DynaThink used in the experiments can be found in Appendix~\ref{baseline}.

\textbf{Backbone Models.} In our experiments, GPT-3.5-turbo, Llama-3.1-8B-Instruct, and Gemini-pro serve as the backbone models. The sampling temperature is set to default values, which is 0.7 for GPT-3.5-turbo \citep{wang2022self}, 0.6 for Llama-3.1-8B-Instruct according to the setting file given by Meta, and 0.9 for Gemini-pro according to the description in Google AI Studio, respectively.

\begin{table*}[t]
  \centering
  \small
\caption{\label{math_example_output}
    Example outputs by FST on two math reasoning datasets, the first of which is on GSM8K dataset and the second of which is on MMLU-pro dataset.
  }
\begin{tabular}{l}
\toprule
\begin{tabular}[c]{@{}p{.96\textwidth}@{}}
There are 5 houses on a street, and each of the first four houses has 3 gnomes in the garden. If there are a total of 20 gnomes on the street, how many gnomes does the fifth house have? \\ \\

\colorbox[rgb]{0.8,0.8,0.8}{\textbf{Output:}} \\
Firstly, we calculate the total number of gnomes in the first four houses. Because the number of gnomes in each of the first four houses is 3, the result is 3 × 4 = 12. \\
Secondly, we calculate the number of gnomes the fifth house has. Because the total number of gnomes is 20, the result is 20 - 12 = 8. \\
Therefore, the answer is 8.
\end{tabular} \\ \bottomrule

\begin{tabular}[c]{@{}p{.96\textwidth}@{}}
What is the value of p in 24 = 2p? \\
(A) p = 4 \ \ \ \ \ \ \ \ \ (B) p = 8\ \ \ \ \ \ \ \ \ (C) p = 12\ \ \ \ \ \ \ \ \ (D) p = 24\\ \\

\colorbox[rgb]{0.8,0.8,0.8}{\textbf{Output:}} \\
Firstly, from 2p = 24, we can know that the value of p is equal to half of 24. Therefore, p = 12. The first step is right. \\
Secondly, the content of (C) is ``p = 12'', which is the same as the result calculated in the previous step. \\
Therefore, the answer is (C).
\end{tabular}          \\ \bottomrule
\end{tabular}
\end{table*}

\subsection{Math Reasoning Task}
\label{mathreasoning}

In this section, we provide the details of the experiments on the math reasoning task, including task description, evaluation metrics, and results. The prompt details of our proposed FST can be found in Appendix~\ref{implementation_specifications_FST}.

\textbf{Task Description.} We conduct experiments on GSM8K dataset \citep{cobbe2021training} and the math sub-dataset in MMLU-pro dataset \citep{wang2024mmlu} to examine the ability of our method in solving math problems. The MMLU-pro dataset greatly increases the complexity of problems of the MMLU \citep{hendrycksmeasuring} dataset.

\textbf{Evaluation Metrics.} By following \citep{wang2023plan, wang2024mmlu}, we evaluate the answers by the metric of \textit{Result Accuracy} (RA), which calculates the percentage of correct results.

\textbf{Results.} Table~\ref{experiment_result} reports the accuracy comparison of our FST and baseline methods on the math reasoning datasets. Our proposed FST achieves the highest accuracy on GSM8K dataset. In addition, due to the increase in problem complexity, the performance improvement of some baseline methods on MMLU-pro dataset over base LLMs is limited. On the contrary, FST still significantly improves the performance of the base LLMs on MMLU-pro dataset, increasing RA by up to 13.85\%. It is worth mentioning that although DynaThink has a slight advantage over our FST on MMLU-pro dataset, it needs to call LLMs hundreds of times for each task on average, and this number increases with the complexity of the task. This is because DynaThink requires that the consistency of multiple outputs from LLMs must meet predefined conditions (more than half of the outputs must be the same, and the reasoning steps for that answer must be minimized), otherwise the number of calls will continue to increase. As the complexity of the task increases, the output of LLMs becomes difficult to maintain high consistency, leading to an increasing number of calls until the preset upper limit (several thousand times) is reached. However, our proposed FST achieves impressive efficiency while achieving high task performance, only requiring calling LLMs three times for each task. Moreover, for each dataset, we provide the output of FST for one problem in Table~\ref{math_example_output} to intuitively demonstrate the performance of FST (see more details in Appendix~\ref{detail_output}).

\subsection{Long-Content Answering Task}
\label{longcontentanswering}

In this section, we provide the details of the experiments on the long-content answering task, including task description, evaluation metrics, and results. The prompt details of our proposed FST can be found in Appendix~\ref{implementation_specifications_FST}.

\textbf{Task Description.} We choose two datasets: 1) LongBench \citep{bai2023longbench}, which requires the models to understand the long content and answer the questions, and 2) the Automated Student Assessment Prize (ASAP) dataset \citep{ASAP}, which requires the models to score the essays according to the specific evaluation protocols.

\begin{table*}[t]
  \centering
  \small
\caption{\label{long_example_output}
    Example outputs by FST on two long-content answering datasets, the first of which is on LongBench dataset and the second of which is on ASAP dataset.
  }
\begin{tabular}{l}
\toprule
\begin{tabular}[c]{@{}p{.96\textwidth}@{}}
There are some paragraphs below. Please determine how many unique paragraphs there are. The output format should only contain the number, such as 1, 2, 3. \\
\textless{}context\textgreater{}\\
The final answer is: \\ \\

\colorbox[rgb]{0.8,0.8,0.8}{\textbf{Output:}} \\
8.
\end{tabular}  \\ \bottomrule

\begin{tabular}[c]{@{}p{.96\textwidth}@{}}
Source Essay: ...\\
Describe the mood created by the author in the memoir. \\
The above is the topic of the article. Now I will give you an article and the evaluation protocols. You need to score the article according to the evaluation protocols.\\ \\
The essay: ...\\
The evaluation protocols of score 0-4: ...\\ \\

\colorbox[rgb]{0.8,0.8,0.8}{\textbf{Output:}} \\
The score of the essay is 3. The response gives a clear and right description of the mood created by the author in the memoir. However, A score of 4 also requires that the response includes relevant and specific information from the memoir. Although the response includes an explanation of it, the explanation is not specific, such as citing content from the memoir and interpreting the author's experience.
\end{tabular}          \\ \bottomrule
\end{tabular}
\end{table*}

\textbf{Evaluation Metrics.} For LongBench dataset, we follow \cite{bai2023longbench} and utilize the \textit{macro-average} (M-Avg), which calculates the average score over major subtask categories. For ASAP dataset, we evaluate the results by the metric of \textit{Quadratic Weighted Kappa} (QWK), which is the evaluation metric of the Kaggle competition on ASAP dataset. Specifically, QWK is used to measure the agreement between two annotators, which is calculated by the following expression:

\begin{equation}
w_{i,j} = \frac{\left ( i- j \right ) ^{2} }{\left ( N- 1 \right ) ^{2} } ,
\end{equation}

\begin{equation}
k= 1- \frac{ {\textstyle \sum_{i,j}^{}  w_{i,j} O_{i,j} } }{ {\textstyle \sum_{i,j}^{}  w_{i,j} E_{i,j} } } ,
\end{equation}

\noindent where $i$ and $j$ denote two scores, and $N$ is the upper bound of the score. Here, $w_{i,j}$ is the weight assigned to the scenario where two raters give scores of $i$ and $j$, respectively, and $O_{i,j}$ is the number of the essays in the dataset that receive two scores of $i$ and $j$. For $E_{i,j}$, it is an element in the $N$-by-$N$ matrix $\textbf{E}$ with size $N$-by-$N$, which is calculated as the outer product between each rater’s histogram vector of ratings, assuming that there is no correlation between rating scores. From these three matrices, the quadratic weighted kappa $k$ is calculated. A larger $k$ corresponds to a higher similarity.

\textbf{Results.} Table~\ref{experiment_result} reports the comparison results of our FST and baseline methods on two long-content answering datasets. Compared with short-content tasks, long-content tasks often contain more restrictions, which pose great difficulty to models. Moreover, we observe that the baseline methods can only achieve the maximal improvement of 6.4\% (M-Avg) on LongBench dataset and 0.173 (QWK) on ASAP dataset. In contrast, our FST increases M-Avg by up to 13.85\% on LongBench dataset, and QWK by up to 0.447 on ASAP dataset. Moreover, for each dataset, we provide the output of FST for one problem in Table~\ref{long_example_output} to intuitively demonstrate the performance of FST (see more details in Appendix~\ref{detail_output}).

\subsection{Constrained Story Generation Task}
\label{constrainedstorygeneration}

In this section, we provide the details of the experiments on the constrained story generation task, including task description, evaluation metrics, and results. The prompt details of our proposed FST can be found in Appendix~\ref{implementation_specifications_FST}.

\textbf{Task Description.} We choose CommonGen-Hard dataset \citep{madaan2024self}, which contains 200 problems, where each problem requires LLMs to generate a story containing 20-30 specified words. Note that it is a quite challenging task compared with other works \citep{lin2020commongen, saha2024branch} that simply require stories to contain no more than 10 specified words.

\textbf{Evaluation Metrics.} We evaluate the stories from two aspects, \textit{i.e.}, constraint satisfaction and overall story quality. For constraint satisfaction, we follow \citep{saha2024branch} and utilize two metrics: 1) \textit{All Present Rate} (APR), which measures the proportion of stories containing all specified words; 2) \textit{Missing Concept Rate} (MCR), which calculates the average percentage of missing specified words. For overall story quality, stories from various methods are ranked based on their qualities through manual evaluation \citep{zheng2024judging}. To ensure the fairness of manual evaluation, a blind review mechanism is used (see Appendix~\ref{human_annotation} for details about the manual assessment process). For each method, we report the proportion of top-ranked stories, which is termed \textit{Optimal Story Ratio} (OSR).

\begin{table*}[t]
  \centering
  \small
\caption{\label{story_example_output}
    Example outputs by FST on constrained story generation dataset. In the answer, the bold and underlined texts denote the words that must be included in the story.
  }
\begin{tabular}{l}
\toprule
\begin{tabular}[c]{@{}p{.96\textwidth}@{}}
Write a short story containing all the following words:\\
vegetable, chess, keyboard, motorcycle, sun, sip, ski, part, shoe, mask, use, hose, racket, paint, wheel, hookah, hole, lawn, wand, base, calf, crash, game. \\ \\

\colorbox[rgb]{0.8,0.8,0.8}{\textbf{Output:}} \\
This is a world full of magic. One day, as the \underline{\textbf{sun}} beat down on the lush \underline{\textbf{lawn}}, a group of friends gathered around an old, abandoned house, wanting to make it their \underline{\textbf{base}}. One of them, Mike arrived on his \underline{\textbf{motorcycle}}. Upon their arrival at the house, they were intrigued by a chessboard set up outside. They need to play \underline{\textbf{chess}}, moving the pieces into special formations to unlock the house. After success, they discovered a basement. In one corner, hidden beneath an old \underline{\textbf{shoe}} and a \underline{\textbf{mask}}, was a \underline{\textbf{hole}} leading into the depths below. They dove into the unknown. They stumbled upon a magic instrument \underline{\textbf{keyboard}} that seemed ancient, a \underline{\textbf{wheel}} etched with peculiar symbols, and a \underline{\textbf{wand}} emitting a faint glow. As they explored, they encountered challenges. They found themselves playing a life-sized \underline{\textbf{game}}, each move determining their fate. With teamwork, they emerged victorious. Finally, it seems they have entered a new world, one where magic doesn't exist. They found that although there was no magic, people \underline{\textbf{used}} tools instead, such as using a \underline{\textbf{hose}} to glue \underline{\textbf{vegetables}}, using a board to \underline{\textbf{ski}}, etc. They are trying new things, such as \underline{\textbf{painting}}, feeding a \underline{\textbf{calf}} grazing, hitting the ball with a \underline{\textbf{racket}}, \underline{\textbf{sipping}} a \underline{\textbf{hookah}}, etc. Suddenly, there was a \underline{\textbf{crash}}. A \underline{\textbf{part}} of the ground opened up. They are sucked into the hole and return to their world, and the previous experience is like a dream.
\end{tabular}  \\ \bottomrule

\end{tabular}
\end{table*}

\textbf{Results.} We report APRs, MCRs, and OSRs of different methods in Table~\ref{experiment_result}. We see that the base models (\textit{i.e.}, GPT-3.5-turbo, Llama-3.1-8B-Instruct, and Gemini-pro) achieve low APRs of no more than 2\%, and high MCRs of no less than 18.00\%, which reveal the complexity and difficulty of this task. It is also notable that the application of the methods, \textit{i.e.}, Zero-Shot-CoT, BoT, SPP, Step-Back Prompting, or BSM, does not necessarily improve the results. Differently, FST significantly increases APR by up to 10\% and reduces MCR by up to 6.25\%, and the results of OSRs indicate that nearly half of the top-ranked stories are generated by FST. It is worth mentioning that although FST is universal, the performance of FST is better than BSM which is designed for constrained story generation tasks. Moreover, we provide the output of FST for one problem in Table~\ref{story_example_output} to intuitively demonstrate the performance of FST (see more details in Appendix~\ref{detail_output}).

\subsection{Performance Investigation}
\label{investigation}

In this section, we conduct in-depth research on our proposed FST to analyze its effectiveness and stability. Considering that Llama-3.1-8B-Instruct is the latest open-sourced LLM, we employ it as the backbone model.

\begin{table}[t]
\centering
  \caption{\label{table4}
    Ablation study on FST. Here, ``$\uparrow$'' means that higher values are better, and ``$\downarrow$'' means that lower values are better. The best records under each metric are highlighted in bold.
  }
\setlength{\tabcolsep}{1.7mm}{
\begin{tabular}{lcccclc}
\hline
\multicolumn{1}{c}{\multirow{2}{*}{Method}} & GSM8K   & MMLU-pro & LongBench & \multicolumn{1}{l}{ASAP} & \multicolumn{2}{c}{CommonGen-Hard} \\ \cmidrule{2-7} 
\multicolumn{1}{c}{}                        & RA $\uparrow$      & RA $\uparrow$       & M-Avg $\uparrow$     & QWK $\uparrow$                      & \multicolumn{1}{c}{APR $\uparrow$}  & MCR $\downarrow$     \\ \hline
Llama-3.1-8B-Instruct                       & 72.50\% & 22.35\%  & 35.64\%   & 0.181                    & 1.00\%                   & 22.39\% \\
\ \ + FT                                        & 63.45\% & 2.66\%   & 10.45\%   & 0.102                    & 0.00\%                   & 71.25\% \\
\ \ + ST                                        & 78.10\% & 23.54\%  & 38.18\%   & 0.296                    & 2.50\%                   & 48.23\% \\
\ \ + FT + ST                                   & 87.35\% & 35.16\%  & 44.84\%   & 0.492                    & 7.00\%                   & 17.22\% \\
\ \ + FT + ST + OI (FST)                                       & \textbf{90.05\%} & \textbf{36.20\%}  & \textbf{46.39\%}   & \textbf{0.583}                    & \textbf{7.50\%}                   & \textbf{16.14\%} \\ \hline
\end{tabular}}
\end{table}

\begin{table}[t]
  \centering
\caption{Stability study on ASAP dataset.}
  \label{table5}
\setlength{\tabcolsep}{3.3mm}{
\begin{tabular}{c|c|c|c|c}
\toprule
Base Model                             & Method               & Disturbance Level & QWK   & Disturbance Degree \\ \midrule
\multirow{4}{*}{Llama-3.1-8B-Instruct} & \multirow{4}{*}{FST} & (No disturbance)  & 0.583 &                    \\
                                       &                      & Character-level   & 0.556 & -0.027             \\
                                       &                      & Word-level        & 0.541 & -0.042             \\
                                       &                      & Semantic-level    & 0.568 & -0.015             \\ \bottomrule
\end{tabular}}
\end{table}

\textbf{Ablation Study.} To show that every step of FST (\textit{i.e.}, FT, ST, and OI) is indispensable, here we conduct an ablation study by comparing the performance of various steps of FST. We investigate the performance on all datasets (\textit{i.e.}, GSM8K, MMLU-pro, LongBench, ASAP, and CommonGen-Hard). We present the results in Table~\ref{table4}. It is worth mentioning that the metric OSR used on the CommonGen-Hand dataset is horizontal (performance comparison between different methods). Our ablation study is longitudinal (performance comparison between different steps of FST), so we did not use the OSR metric. As shown in Table~\ref{table4}, the performance only using FT is poor, and the improvement brought by only using ST is marginal. However, the performance of the base LLM is significantly improved by the combination of FT and ST, such as increasing RA by up to 14.85\% on GSM8K dataset, M-Avg by up to 9.20\% on LongBench dataset, and APR by up to 6.00\% on CommonGen-Hard dataset. Moreover, the OI step further improves the performance of the base LLM under FT + ST through reflection and refinement. The results demonstrate the effectiveness of combining fast thinking and slow thinking, as well as the importance of each step of FST.

\textbf{Stability Study.} The performance stability brought by the LLM-based methods under slight disturbances in the prompts is an important issue. Therefore, we conduct the experiments to test the performance stability of our proposed FST. Inspired by \citep{zhu2023promptbench}, we introduce three levels of disturbances into the prompt of FST: 1) \textit{character-level}, which manipulates texts by introducing typos or errors to words, e.g., by adding, deleting, repeating, replacing, and permuting characters for certain words; 2) \textit{word-level}, which manipulates texts by replacing some words with synonyms or contextually similar words; 3) \textit{semantic-level}, which manipulates texts by rewriting some sentences while ensuring that the semantic information of these sentences remains unchanged. The specific implementations of them can be found in Appendix~\ref{level_stability_study}. Using ASAP as the experimental dataset, we provide the experimental results in Table~\ref{table5}. As shown in Table~\ref{table5}, the performance of FST with disturbances shows little decrease, with QWK only decreasing by a maximum of 0.042. This is because our designed prompts can uncover the potential ability of LLMs in FT and ST. As a result, as long as the semantic information of prompts remains unchanged, small linguistic disturbances to the prompt will not have too much impact on the effectiveness of FST.




\section{Conclusion}

The performance of popular LLMs may be suboptimal when facing increasingly difficult tasks that involve overly complex logic and numerous constraints. To deal with this issue, we are inspired by the human cognition process and proposed a new task decomposition algorithm dubbed ``Fast-Slow-Thinking’’ (FST) in this paper, which can enhance the performance of LLMs on various complex tasks such as math reasoning, long-content answering, and constrained story generation. We believe that mimicking human thinking modes can unlock the potential of LLMs, and we hope that our work will inspire further research on integrating the ways of human thinking into the development of LLMs.

\backmatter





\begin{appendices}

\section{Implementation Specifications of FST}
\label{implementation_specifications_FST}

We provide the specific prompt template of our proposed FST in Table~\ref{prompt_template_FST}. The template contains three steps: Fast Thinking, Slow Thinking, and Output Inspection. Every sentence is carefully designed in each step to make LLMs follow the specifications and generate high-quality responses.

Based on this prompt template, we provide the specific prompts of FST for three types of tasks mentioned in Section~\ref{experiments} in Tables~\ref{table6}, \ref{table8}, and \ref{table10}.

\section{Implementations of Baseline Methods}
\label{baseline}

\begin{figure*}[t]
  \centering
  \includegraphics[width=\textwidth, height=1.4\columnwidth]{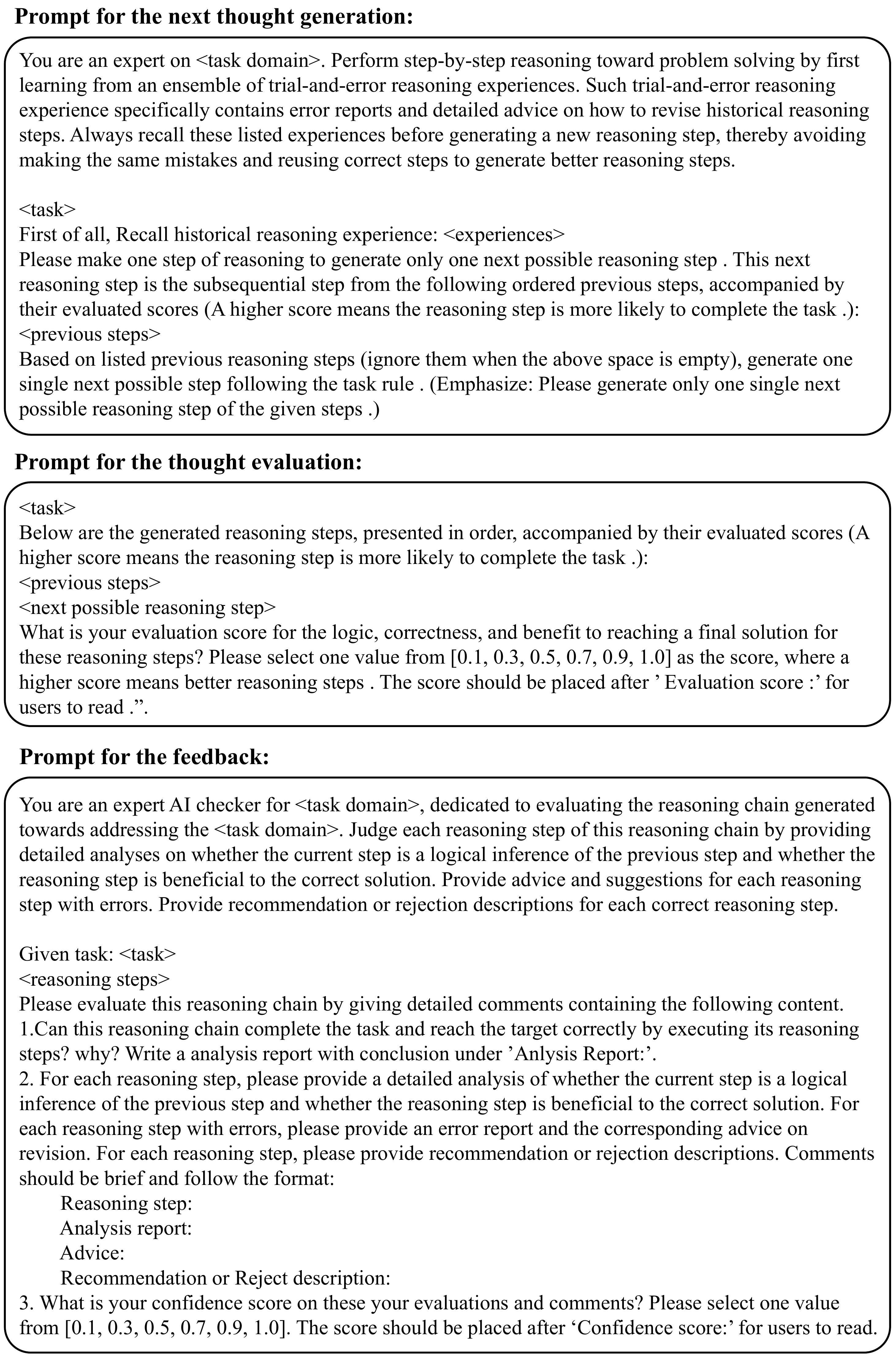}
  \caption {Prompt of Boosting of Thoughts \citep{chen2024boosting} used in the experiments.}
  \label{BoT}
\end{figure*}

\begin{figure*}[t]
  \centering
  \includegraphics[width=\textwidth, height=1.4\columnwidth]{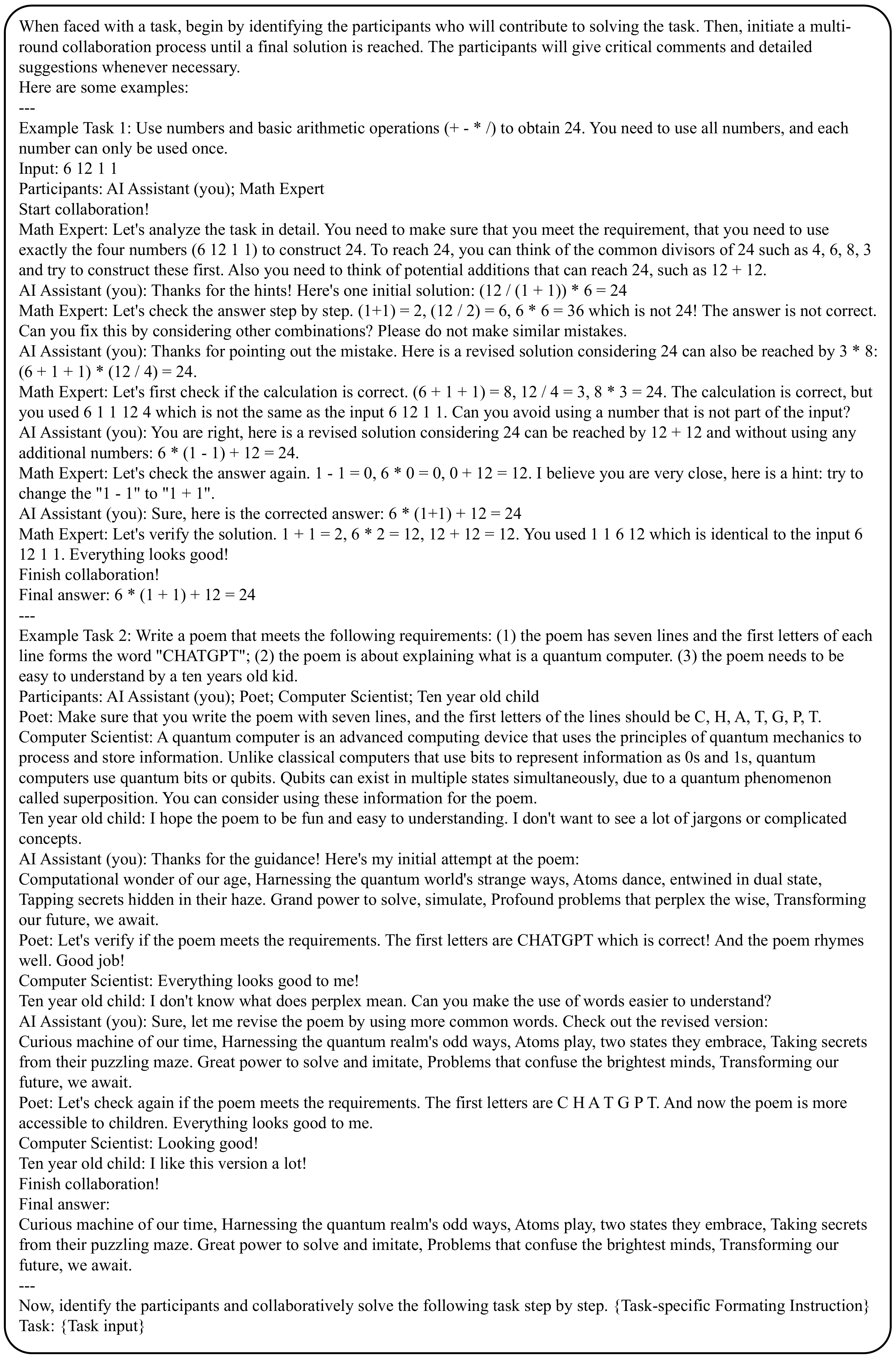}
  \caption {Prompt of Solo Performance Prompting \citep{wang2024unleashing} used in the experiments.}
  \label{SPP}
\end{figure*}

\begin{figure*}[t]
  \centering
  \includegraphics[width=\textwidth, height=1.4\columnwidth]{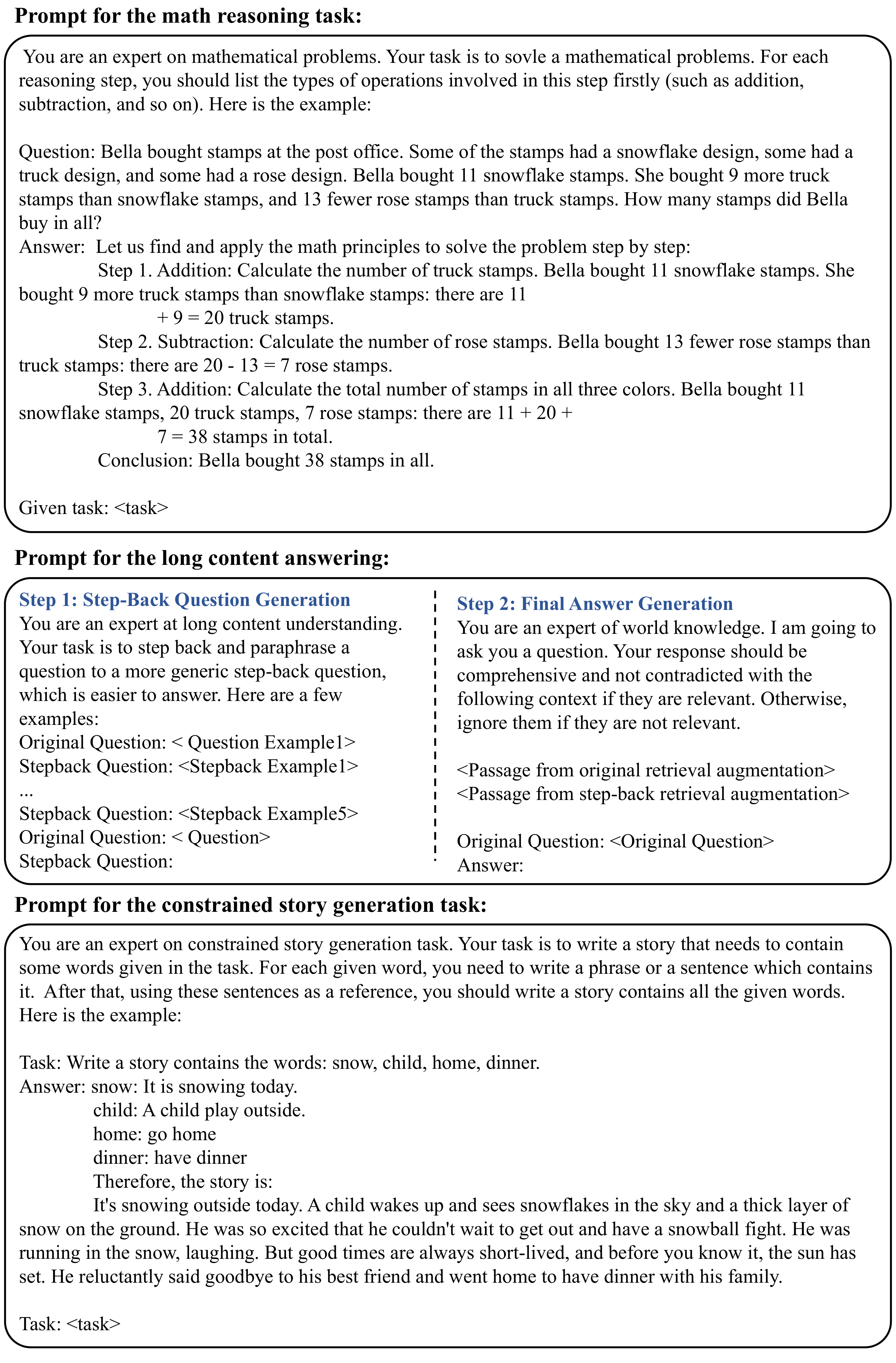}
  \caption {Prompt of Step-Back Prompting \citep{zheng2024take} used in the experiments.}
  \label{SBP}
\end{figure*}

\begin{figure*}[t]
  \centering
  \includegraphics[width=\linewidth]{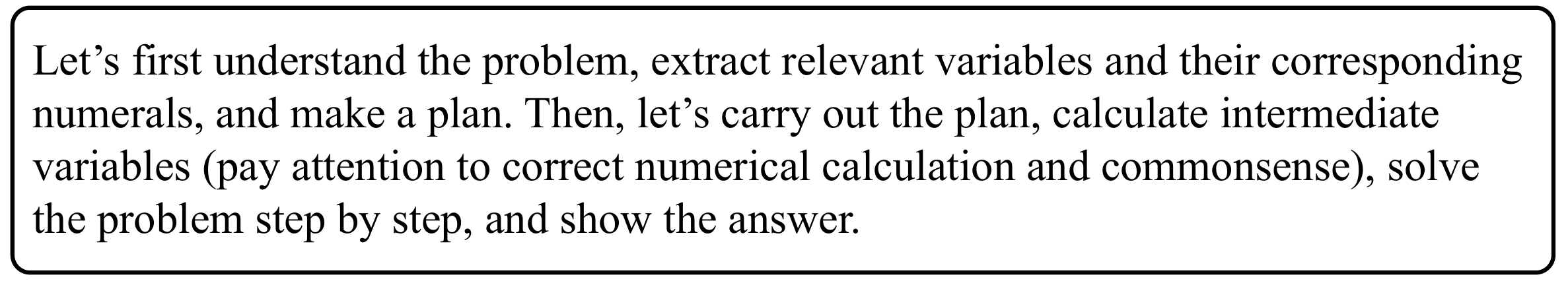}
  \caption {Prompt of Plan-and-Solve Method \citep{wang2023plan} used in the experiments.}
  \label{PS}
\end{figure*}

\begin{figure*}[t]
  \centering
  \includegraphics[width=\linewidth]{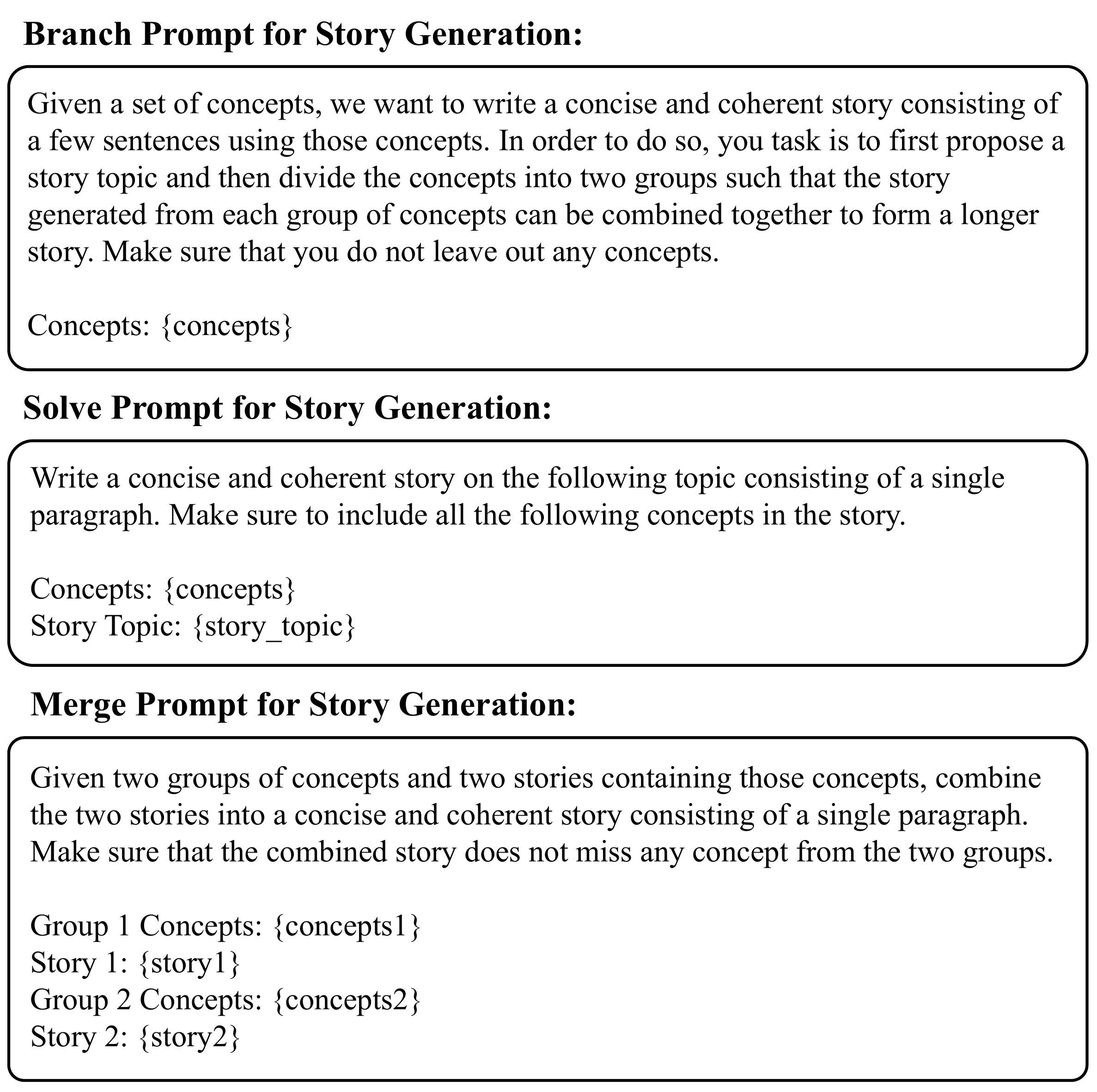}
  \caption {Prompt of Branch-Solve-Merge Method \citep{saha2024branch} used in the experiments.}
  \label{BSM}
\end{figure*}

As shown in Figures~\ref{BoT}, ~\ref{SPP}, ~\ref{SBP}, ~\ref{PS}, and ~\ref{BSM}, we provide the specific prompts of the baseline methods (\textit{i.e.}, Boosting of Thoughts, Solo Performance Prompting, Step-Back Prompting, Plan-and-Solve, and Branch-Solve-Merge) used in our experiments. As shown in Algorithm~\ref{DynaThink}, we provide the implementation details of DynaThink. For all the baseline methods, the experimental settings are the same as those in their papers, respectively.

\section{Details of Manual Assessment Process}
\label{human_annotation}

To ensure the objectivity and fairness of the manual evaluation results and also accurately reflect the quality of the answers, the manual assessment process is carefully designed, as described below.

\textbf{Volunteers.} A total of 7 volunteers participate in the manual assessment process, all of whom are undergraduates from well-known universities. They have the right values and aesthetics. Moreover, before the formal manual evaluation process, these volunteers have received relevant training and passed the tests, so they possess the ability to evaluate stories correctly.

\textbf{Data Preprocessing.} To ensure the fairness of the manual assessment, we adopt the blind review mechanism, where the order of the answers generated by different methods for the same task is randomized. Moreover, we carefully design the questions of the manual assessment experiment so that the volunteers can clearly understand the needs of the experiment and give objective assessments.

The specific templates are as follows: ``Six stories are shown below. Please read the content of each story carefully, and comprehensively consider many aspects (story theme, contextual logic, sentence fluency, plot completeness, \textit{etc.}). After that, you need to give a ranking of 6 stories, of which 1 is the best, and 6 is the worst. \textless six stories\textgreater.''

\textbf{Assessment Process.} Each set of stories (six stories generated by different methods for the same task) is given to all volunteers, resulting in 7 ranking results. During the assessment process, volunteers do not discuss with each other and carry out the assessment independently.

\section{Stability Study}
\label{level_stability_study}

To compare the performance stability of our proposed FST, we add disturbances to the prompt of FST from three levels, \textit{i.e.}, character-level, word-level, and semantic-level. In this section, we provide their implementations in our experiments for stability study.

\textbf{Character-level.} In our experiments, as shown in Table~\ref{table17}, we add two syntax errors and word spelling errors to the prompt of each step in FST (\textit{i.e.}, FT, ST, and OI).

\textbf{Word-level.} In our experiments, as shown in Table~\ref{table18}, we replace four words with synonyms or contextually similar words in the prompt of each step of FST (\textit{i.e.}, FT, ST, and OI).

\textbf{Semantic-level.} In our experiments, as shown in Table~\ref{table20}, we rewrite three sentences in the prompt of each step of FST (\textit{i.e.}, FT, ST, and OI) while ensuring that their semantics remain unchanged.

\section{Examples of FST Outputs}
\label{detail_output}

To visually demonstrate the effectiveness of our proposed FST, we provide example outputs generated by FST for each dataset mentioned in Section~\ref{experiments}. As shown in Tables~\ref{table12}, \ref{table13}, \ref{table14}, \ref{table15}, and \ref{table16}, for each example, we list the outputs of each step in FST.

\begin{table*}[t]
\caption{\label{table6}
    Specific Prompt of FST used on the math reasoning task.
  }
  \centering
  \small
\resizebox{\columnwidth}{!}{
\begin{tabular}{cl}
\toprule
Step                   & \multicolumn{1}{c}{Prompt Template}                                           
\\ \midrule
FT                         & \begin{tabular}[c]{@{}p{\textwidth}@{}}\\You are an expert in math reasoning. You are very good at understanding and solving tasks in this domain. Now I will give you a complex task, \textless{}the task\textgreater{}. Your work should follow the two steps: \\ \\ Step 1: You need to understand the task and simplify the task into a concise and general one.\\ I give you some simplification examples below as guidance:\\   \\      Task 1: Tom had three apples. He ate one and gave one to Jane. How many apples he has now? \\ Simplification Task 1: Which data is related to the number of apples Tom has now?\\ Task 2: Tom can read one page in five minutes. Today he needs to read 120 pages. How many hours will it take him to finish? \\ Simplification Task 2: Which data is related to the time that Tom needs to take? \\ \\ Step 2: Please generate the answer to the concise and general task.\\ \\ \end{tabular}                                                                                                                                                                                                                                                                                                   \\  \midrule
ST                         & \begin{tabular}[c]{@{}p{\textwidth}@{}}\\ Based on the concise and general task \textless{}the concise and general task\textgreater{}, I will add some constraints:\\ \textless{}the constraints removed in the Fast Thinking step\textgreater{}\\ \\ Please take a deep consideration of these constraints and improve the answer of the concise and general task below to meet these constraints. \\ \textless the answer of the concise and general task \textgreater \\ \\ Tips: \\ 1. Pay attention to the correctness of the intermediate process.\\ 2. The logic must be reasonable.\\ \end{tabular} \\ \\  \midrule

OI & \begin{tabular}[c]{@{}p{\textwidth}@{}}\\ \textless{}the original task\textgreater\\ The answer is: \textless{}the answer generated in the Slow Thinking step\textgreater\\ \\ You need to check the answer through the following steps:\\ Step 1: Whether the answer strictly meets the requirements of the task. If not, please improve it.\\ Step 2: Whether the intermediate process is correct. If not, please improve it.\\ Step 3: Can every sentence of the answer be supported by the problem and material given to you? If not, modify unsupported parts.\\ \\ \end{tabular}                                                           \\ \bottomrule
\end{tabular}}
\end{table*}

\begin{table*}[t]
\caption{\label{table8}
    Specific Prompt of FST used on the long-content answering task.
  }
  \centering
\small
\resizebox{\columnwidth}{!}{
\begin{tabular}{cl}
\toprule
Step                   & \multicolumn{1}{c}{Prompt Template}                                           
\\ \midrule
FT                         & \begin{tabular}[c]{@{}p{\textwidth}@{}}\\You are an expert in long-content answering. You are very good at understanding and solving tasks in this domain. Now I will give you a complex task, \textless{}the task\textgreater{}. Your work should follow the two steps: \\ \\ Step 1: You need to understand the task and simplify the task into a concise and general one.\\ I give you some simplification examples below as guidance:\\   \\      Task 1: Summarize this article.\\ Simplification Task 1: Summarize the content of each paragraph in the article.\\ Task 2: \textless{}The essay\textgreater{} \\ \ \ \ \ \ \ \ \ \ \ \ Score the essay above according to the following evaluation protocols.\\ Simplification Task 2: Regardless of the protocols, evaluate the essay and write a comment. \\ \\ Step 2: Please generate the answer to the concise and general task.\\ \\ \end{tabular}                                                                                                                                                                                                                                                                                                   \\  \midrule
ST                         & \begin{tabular}[c]{@{}p{\textwidth}@{}}\\ Based on the concise and general task \textless{}the concise and general task\textgreater{}, I will add some constraints:\\ \textless{}the constraints removed in the Fast Thinking step\textgreater{}\\ \\ Please take a deep consideration of these constraints and improve the answer of the concise and general task below to meet these constraints. \\ \textless the answer of the concise and general task \textgreater \\ \\ Tips: \\ 1. Pay attention to the correctness of the intermediate process.\\ 2. The logic must be reasonable.\end{tabular} \\ \\  \midrule

OI & \begin{tabular}[c]{@{}p{\textwidth}@{}}\\ \textless{}the original task\textgreater\\ The answer is: \textless{}the answer generated in the Slow Thinking step\textgreater\\  \\You need to check the answer through the following steps:\\ Step 1: Whether the answer strictly meets the requirements of the task. If not, please improve it.\\ Step 2: Whether the intermediate process is correct. If not, please improve it.\\ Step 3: Can every sentence of the answer be supported by the problem and material given to you? If not, modify unsupported parts.\\ \\ \end{tabular}                                                           \\ \bottomrule
\end{tabular}}
\end{table*}

\begin{table*}[t]
\caption{\label{table10}
    Specific Prompt of FST used on the constrained story generation task.
  }
  \centering
  \small
\resizebox{\columnwidth}{!}{
\begin{tabular}{cl}
\toprule
Step                   & \multicolumn{1}{c}{Prompt Template}                                           
\\ \midrule
FT                         & \begin{tabular}[c]{@{}p{\textwidth}@{}}\\You are an expert in constrained story generation. You are very good at understanding and solving tasks in this domain. Now I will give you a complex task, \textless{}the task\textgreater{}. Your work should follow the two steps: \\ \\Step 1: You need to understand the task and simplify the task into a concise and general one.\\ I give you some simplification examples below as guidance:\\   \\      Task 1: Write a story containing all the following words: snow, child, sky, cloud, tree, happy.\\ Simplification Task 1: Generate a story belonging to a topic that is relevant to the specified words. \\ \\ Step 2: Please generate the answer to the concise and general task.\\ \\ \end{tabular}                                                                                                                                                                                                                                                                                                   \\  \midrule
ST                         & \begin{tabular}[c]{@{}p{\textwidth}@{}}\\ Based on the concise and general task \textless{}the concise and general task\textgreater{}, I will add some constraints:\\ \textless{}the constraints removed in the Fast Thinking step\textgreater{}\\ \\Please take a deep consideration of these constraints and improve the answer of the concise and general task below to meet these constraints. \\ \textless the answer of the concise and general task \textgreater \\ \\ Tips: \\ 1. Pay attention to the correctness of the intermediate process.\\ 2. The logic must be reasonable.\end{tabular} \\ \\  \midrule

OI & \begin{tabular}[c]{@{}p{\textwidth}@{}}\\ \textless{}the original task\textgreater\\ The answer is: \textless{}the answer generated in the Slow Thinking step\textgreater\\ \\You need to check the answer through the following steps:\\ Step 1: Whether the answer strictly meets the requirements of the task. If not, please improve it.\\ Step 2: Whether the intermediate process is correct. If not, please improve it.\\ Step 3: Can every sentence of the answer be supported by the problem and material given to you? If not, modify unsupported parts.\\ \\ \end{tabular}                                                           \\ \bottomrule
\end{tabular}}
\end{table*}

\begin{algorithm}
\caption{DynaThink}\label{DynaThink}
\begin{algorithmic}[1]
\Require Problem set, $P$
\Ensure Fast thinking question set, $Q_{f}$ and Slow thinking question set, $Q_{s}$ 
\State $Q_{f} \gets \varnothing$, $Q_{s} \gets \varnothing$ $\triangleright$ Initialize set for fast questions and slow questions
$n \gets 2$ $\triangleright$ n can be initialized by any integer less than total generation times
\Repeat
\State Generate $n$ responses for problem set $P$ by querying the LLM; store as $Q$.
\State Initialize question set $Q_{1}$, $Q_{2}$ and $Q_{3}$ as $\varnothing$.
\State Calculate voting distribution $F$ for each question in $Q$ based on consistency.
\For{each question $i$ in $Q$}
\State Determine the answer with the highest votes, $a_{j}$, and its vote count, $\max (F(i))$
\If{$\max (F(i))\ge \left \lfloor \frac{n}{2}  \right \rfloor +  1$}
\State Add $Q(i)$ to $Q_{1}$ $\triangleright$ First selected question set
\Else
\State Add $Q(i)$ to $Q_{2}$ $\triangleright$ Set for slow questions
\EndIf
\EndFor
\For{each question in $Q_{1}$}
\State Extract the minimum step array from each answer and determine the step distribution.
\For{each question $i$}
\State Find the minimum steps, $\min(Steps(i))$, and
the step of the majority-voted answer, $a_{i}$.
\If{$a_{i}==\min(Steps(i))$}
\State Add $Q_{1}(i)$ to $Q_{3}$
\Else
\State Add $Q_{1}(i)$ to $Q_{2}$
\EndIf
\EndFor
\EndFor
\If{$Q_{3} \ne \varnothing$}
\State Update $Q_{f}$ with $Q_{f} \cup Q_{3}$
\State Update $P$ with the questions in $Q_{2}$
\State increase $n$ $\triangleright$ n can increase by any integer based on the budget limit
\Else
\State $Q_{s} = Q_{2}$
\EndIf
\Until{$Q_{3} = \varnothing$}
\State return $Q_{f}$, $Q_{s}$
\end{algorithmic}
\end{algorithm}

\begin{table*}[t]
  \centering
  \small
\caption{\label{table17}
    Character-level disturbances of FST. The \textbf{texts highlighted in bold} denote the specific disturbances we add in the prompt of FST. 
  }
\resizebox{\columnwidth}{!}{
\begin{tabular}{cl}
\toprule
Step                   & \multicolumn{1}{c}{Prompt Template}                                           
\\ \midrule
FT                         & \begin{tabular}[c]{@{}p{\textwidth}@{}}\\You \textbf{is} an expert in long-content \textbf{answer}. You are very good at \textbf{understand} and solving tasks in this domain. Now I will give you a complex task, \textless{}the task\textgreater{}. Your work should follow the two steps: \\ \\Step 1: You need to understand the task and \textbf{samplify} the task into a concise and general one.\\ I give you some simplification examples below as guidance:\\   \\      Task 1: \textless{}The essay\textgreater{} \\ \ \ \ \ \ \ \ \ \ \ \ Score the essay above according to the following evaluation protocols\\ Simplification Task 1: Regardless of the protocols, evaluate the essay and write a comment. \\ \\ Step 2: Please \textbf{generade} the answer of the concise and general task.\\ \\ \end{tabular}                                                                                                                                                                                                                                                                                                   \\  \midrule
ST                         & \begin{tabular}[c]{@{}p{\textwidth}@{}}\\ Based on the concise and general task \textless{}the concise and general task\textgreater{}, I will \textbf{adds} some constraints:\\ \textless{}the constraints removed in the Fast Thinking step\textgreater{}\\ \\Please take a deep \textbf{considerasion} of these constraints and \textbf{improving} the answer of the concise and general task below to meet these constraints. \\\textless the answer of the concise and general task\textgreater \\ \\ Tips: \\ 1. Pay attention to the correctness of the intermediate process.\\ 2. The logic must be reasonable.\end{tabular} \\ \\  \midrule

OI & \begin{tabular}[c]{@{}p{\textwidth}@{}}\\ \textless{}the original task\textgreater\\ The answer \textbf{are}: \textless{}the answer generated in the Slow Thinking step\textgreater\\ \\ You need to check the answer \textbf{thtough} the following steps:\\ Step 1: Whether the answer strictly meets the requirements of the task. If not, please improve it.\\ Step 2: Whether the \textbf{intetnediate} process is correct. If not, please improve it.\\ Step 3: Can every sentence of the answer be supported by the problem and material given to you? If not, modify unsupported parts.\\ \\ \end{tabular}                                                           \\ \bottomrule
\end{tabular}}
\end{table*}

\begin{table*}[t]
  \centering
  \small
\caption{\label{table18}
    Word-level disturbances of FST. The \textbf{texts highlighted in bold} denote the specific disturbances we add in the prompt of FST.
  }
\resizebox{\columnwidth}{!}{
\begin{tabular}{cl}
\toprule
Step                   & \multicolumn{1}{c}{Prompt Template}                                           
\\ \midrule
FT                         & \begin{tabular}[c]{@{}p{\textwidth}@{}}\\You are an expert in long-content \textbf{understanding}. You are very good at understanding and solving tasks in this domain. Now I will give you a complex task, \textless{}the task\textgreater{}. Your work should follow the two steps: \\ \\Step 1: You need to understand the task and simplify the task into a concise and general one.\\ I give you some simplification \textbf{cases} below as guidance:\\   \\      Task 1: \textless{}The essay\textgreater{} \\ \ \ \ \ \ \ \ \ \ \ \ Score the essay above according to the following evaluation protocols\\ Simplification Task 1: Regardless of the \textbf{metrics}, evaluate the essay and write a comment. \\ \\ Step 2: Please generate the \textbf{response} of the concise and general task.\\ \\ \end{tabular}                                                                                                                                                                                                                                                                                                   \\  \midrule
ST                         & \begin{tabular}[c]{@{}p{\textwidth}@{}}\\ Based on the concise and general task \textless{}the concise and general task\textgreater{}, I will add some \textbf{requirements}:\\ \textless{}the constraints removed in the Fast Thinking step\textgreater{}\\ \\Please take a deep consideration of these constraints and improve the answer of the concise and general task below to \textbf{satisfy} these constraints. \\\textless the answer of the concise and general task\textgreater \\ \\ Tips: \\ 1. Pay attention to the correctness of the intermediate \textbf{reasoning}.\\ 2. The logic must be reasonable.\end{tabular} \\ \\  \midrule

OI & \begin{tabular}[c]{@{}p{\textwidth}@{}}\\ \textless{}the original task\textgreater\\ The \textbf{result} is: \textless{}the answer generated in the Slow Thinking step\textgreater\\ \\ You need to \textbf{review} the answer through the following steps:\\ Step 1: Whether the answer strictly meets the requirements of the task. If not, please improve it.\\ Step 2: Whether the intermediate process is correct. If not, please improve it.\\ Step 3: Can every sentence of the answer be supported by the \textbf{context} given to you? If not, modify unsupported parts.\\ \\ \end{tabular}                                                           \\ \bottomrule
\end{tabular}}
\end{table*}

\begin{table*}[t]
  \centering
  \small
\caption{\label{table20}
    Semantic-level disturbances of FST. The \textbf{texts highlighted in bold} denote the specific disturbances we add in the prompt of FST.
  }
\resizebox{\columnwidth}{!}{
\begin{tabular}{cl}
\toprule
Step                   & \multicolumn{1}{c}{Prompt Template}                                           
\\ \midrule
FT                         & \begin{tabular}[c]{@{}p{\textwidth}@{}}\\ \textbf{You are professional in answering long-content questions.} You are very good at understanding and solving tasks in this domain. \textbf{Now you face a task, \textless{}the task\textgreater{}. and need to answer follow two steps:} \\ \\Step 1: You need to understand the task and simplify the task into a concise and general one.\\ I give you some simplification examples below as guidance:\\   \\      Task 1: \textless{}The essay\textgreater{} \\ \ \ \ \ \ \ \ \ \ \ \ Score the essay above according to the following evaluation protocols. \\ Simplified Task 1: Regardless of the protocols, evaluate the essay and write a comment. \\ \\ \textbf{Step 2: Please complete this concise and general task.}\\ \\ \end{tabular}                                                                                                                                                                                                                                                                                                   \\  \midrule
ST                         & \begin{tabular}[c]{@{}p{\textwidth}@{}}\\ \textbf{On the basis of this task:} \textless{}the concise and general task\textgreater{}, I will add some constraints:\\ \textless{}the constraints removed in the Fast Thinking step\textgreater{}\\ \\  \textbf{Please consider these constraints deeply and improve the solution below to satisfy these constraints.} \\\textless the answer of the concise and general task\textgreater \\ \\ Tips: \\ \textbf{1. The intermediate process needs to be correct.}\\ 2. The logic must be reasonable.\end{tabular} \\ \\  \midrule

OI & \begin{tabular}[c]{@{}p{\textwidth}@{}}\\ \textless{}the original task\textgreater\\ The answer is: \textless{}the answer generated in the Slow Thinking step\textgreater\\ \\ \textbf{Please check the response from the following two aspects:}\\ \textbf{Step 1: If the answer does not meet the requirements, improve it.}\\ Step 2: Whether the intermediate process is correct. If not, please improve it.\\ \textbf{Step 3: Can the content of the answer be supported by the provided context? If not, modify unsupported parts.}\\ \\ \end{tabular}                                                           \\ \bottomrule
\end{tabular}}
\end{table*}

\begin{table*}[t]
  \centering
  \footnotesize
\caption{\label{table12}
    Example outputs by FST for GSM8K dataset. The \textbf{texts highlighted in bold} denote the key intermediate process generated by FST to solve this task.
  }
\resizebox{\columnwidth}{!}{
\begin{tabular}{l}
\toprule
\begin{tabular}[c]{@{}p{.96\textwidth}@{}}
\\
\colorbox[rgb]{0.8,0.8,0.8}{\textbf{Question:}} \\
There are 5 houses on a street, and each of the first four houses has 3 gnomes in the garden. If there are a total of 20 gnomes on the street, how many gnomes does the fifth house have? \\ \\
\colorbox[rgb]{0.8,0.8,0.8}{\textbf{Answer (FT):}} \\
Simplification Task: \textbf{Which data is related to the number of gnomes the fifth house has?} \\
Answer: The total number of gnomes, the number of gnomes in the first four houses. \\ \\

\colorbox[rgb]{0.8,0.8,0.8}{\textbf{Answer (ST):}}\\
To know the number of gnomes the fifth house has, we need to know the total number of gnomes and the number of gnomes in the first four houses. \\
\textbf{We can know that the total number of gnomes is 20 and the number of gnomes in each of the first four houses is 3 from the constraints.} Therefore, the total number of gnomes in the first four houses is 3 × 4 = 12. \\
According to the results above, the number of gnomes the fifth house has is 20 - 12 = 8. \\
Therefore, the result of the question is 8. \\ \\

\colorbox[rgb]{0.8,0.8,0.8}{\textbf{Answer (OI):}} \\
Let us check the answer: \\
\textbf{Firstly}, we calculate the total number of gnomes in the first four houses. Because the number of gnomes in each of the first four houses is 3, the result is 3 × 4 = 12. \\
\textbf{Secondly}, we calculate the number of gnomes the fifth house has. Because the total number of gnomes is 20, the result is 20 - 12 = 8. \\
Therefore, the answer generated in ST is correct. The result is 8.
\end{tabular} \\ \\ \bottomrule

\begin{tabular}[c]{@{}p{.96\textwidth}@{}}
\\
\colorbox[rgb]{0.8,0.8,0.8}{\textbf{Question:}} \\
On a school trip to the seashore, Alan and his friends collected shells. Alan collected four times as many shells as Ben did. Ben got a late start and only collected a third of what Laurie did. If Laurie collected 36 shells how many did Alan collect? \\ \\
\colorbox[rgb]{0.8,0.8,0.8}{\textbf{Answer (FT):}} \\
Simplification Task: \textbf{Which data is related to the number of shells Alan collected?} \\
Answer: The number of shells Ben collected. \\ \\

\colorbox[rgb]{0.8,0.8,0.8}{\textbf{Answer (ST):}}\\
To know the number of shells Ben collected, we need to know the number of shells Laurie collected. \\
\textbf{We can find that Laurie collected 36 shells according to the new constraints given in the task.} Therefore, one-third of 36 is 12, which is the number of shells Ben collected. \\
Knowing the results above, the number of shells Alan collected is 4 × 12 = 48. \\
Therefore, the result is 48. \\ \\

\colorbox[rgb]{0.8,0.8,0.8}{\textbf{Answer (OI):}} \\
Let us check the answer: \\
\textbf{“one-third of 36 is 36 / 3 = 12” is correct and “four times of 12 is 4 × 12 = 48” is correct.} \\
Therefore, the answer is correct.
\end{tabular}         \\ \\ \bottomrule
\end{tabular}}
\end{table*}

\begin{table*}[t]
  \centering
  \footnotesize
\caption{\label{table13}
    Example outputs by FST for MMLU-pro dataset. The \textbf{texts highlighted in bold} denote the key intermediate process generated by FST to solve this task.
  }
\begin{tabular}{l}
\toprule
\begin{tabular}[c]{@{}p{.96\textwidth}@{}}
\\
\colorbox[rgb]{0.8,0.8,0.8}{\textbf{Question:}} \\
What is the value of p in 24 = 2p? \\
(A) p = 4 \ \ \ \ \ \ \ \ \ (B) p = 8\ \ \ \ \ \ \ \ \ (C) p = 12\ \ \ \ \ \ \ \ \ (D) p = 24\\ \\
\colorbox[rgb]{0.8,0.8,0.8}{\textbf{Answer (FT):}} \\
Simplification Task: \textbf{Which data is related to the choice of the task?} \\
Answer: The value of half of 24 and the content of each choice. \\ \\

\colorbox[rgb]{0.8,0.8,0.8}{\textbf{Answer (ST):}}\\
Firstly, the value of half of 24 is 24 / 2 = 12.\\
Secondly, the content of (A) is 4, the content of (B) is 8, the content of (C) is 12, and the content of (D) is 24.\\
\textbf{Thirdly, the value of p is 12, which is the same as the content of (C).}\\
Therefore, the result of the question is (C). \\ \\

\colorbox[rgb]{0.8,0.8,0.8}{\textbf{Answer (OI):}} \\
Let us check the answer: \\
\textbf{Firstly, from 2p = 24, we can know that the value of p is equal to half of 24.} Therefore, p = 12. The first step is right. \\
Secondly, the content of (C) is ``p = 12'', which is the same as the result calculated in the previous step. \\
Therefore, the answer generated in ST is correct. The result is (C).
\end{tabular} \\ \\ \bottomrule

\begin{tabular}[c]{@{}p{.96\textwidth}@{}}
\\
\colorbox[rgb]{0.8,0.8,0.8}{\textbf{Question:}} \\
If 3 cans of pears cost \$2.37 how many cans of pears can you buy for \$9.48?\\
(A) 3 cans \ \ \ \ \ \ \ \ \ (B) 12 cans\ \ \ \ \ \ \ \ \ (C) 36 cans\ \ \ \ \ \ \ \ \ (D) 13 cans\\ \\
\colorbox[rgb]{0.8,0.8,0.8}{\textbf{Answer (FT):}} \\
Simplification Task: \textbf{Which data is related to the choice of the task?} \\
Answer: The price of a can of pears, the value of \$9.48 divided by the unit price of the product, and the content of each choice. \\ \\

\colorbox[rgb]{0.8,0.8,0.8}{\textbf{Answer (ST):}}\\
Firstly, \textbf{assuming the price of a can of pears is x, we can know that 3x = 2.37 from the constraints.} Then x = 2.37 / 3 = 0.79. Therefore, the price of a can of pears is \$0.79. \\
Secondly, the value of 9.48 / 0.79 is 12. It means that you can buy 12 cans of pears with \$9.48. \\
Thirdly, \textbf{according to the contents of the four choices, we can find that the content of (D) is the closest to the answer to the problem.}
Therefore, the result is (D). \\ \\

\colorbox[rgb]{0.8,0.8,0.8}{\textbf{Answer (OI):}} \\
Let us follow your requirements and check the answer:\\
Step 1: The result is one option, which meets the requirements for multiple-choice questions.\\
Step 2: Check the calculation. ``2.37 / 3 = 0.79'' and ``9.48 / 0.79 = 12'' are correct. \\
\textbf{Step 3: The content of (D) is ``13 cans'', which is different from the calculation result. However, the content of (B) is the same as the calculation result.}\\
Therefore, the answer is wrong. The correct choice is (B).
\end{tabular}         \\ \\ \bottomrule
\end{tabular}
\end{table*}

\begin{table*}[t]
  \centering
  \small
\caption{\label{table14}
    Example outputs by FST for LongBench dataset. The \textbf{texts highlighted in bold} denote the key intermediate process generated by FST to solve this task. It is worth mentioning that the answer generated in ST may contain the intermediate process which violates the constraints ``The output format should only contain the number.'' However, the review in OI is expected to correct this error and generate an answer containing only a number.
  }
\begin{tabular}{l}
\toprule
\begin{tabular}[c]{@{}p{.96\textwidth}@{}}
\\
\colorbox[rgb]{0.8,0.8,0.8}{\textbf{Question:}} \\
There are some paragraphs below sourced from Wikipedia. Some of them may be duplicates. Please carefully read these paragraphs and determine how many unique paragraphs there are after removing duplicates. In other words, how many non-repeating paragraphs are there in total? \\
\textless{}context\textgreater{}\\
Please enter the final count of unique paragraphs after removing duplicates. The output format should only contain the number, such as 1, 2, 3, and so on.\\ \\
The final answer is: \\ \\

\colorbox[rgb]{0.8,0.8,0.8}{\textbf{Answer (FT):}} \\
Simplification Task: \textbf{Which paragraphs are repeated among all the given paragraphs?} \\
Answer: \\
There are many repeated paragraphs in the given 40 paragraphs after checking the contents of them carefully. \textbf{Below are 8 paragraph sets.} In each set, the contents of the paragraphs are the same.\\
(1) Paragraph 1, 11, 12, 15, 27, 31, and 36.\\
(2) Paragraph 2, 6, 18, 30, 33, and 38.\\
(3)Paragraph 3, 7, 10, 20, 25, and 35.\\
(4) Paragraph 4, 14, 22, 29, 32, and 39.\\
(5) Paragraph 5, 8, and 21.\\
(6) Paragraph 9, 13, 23, 34, and 40.\\
(7) Paragraph 16, 24, 26, 28, and 37.\\
(8) Paragraph 17, and 19.\\ \\

\colorbox[rgb]{0.8,0.8,0.8}{\textbf{Answer (ST):}}\\
Firstly, the given answer is the correct answer to the simplification task. According to the paragraph numbers given by each group, I compare the paragraph contents and find that the paragraphs in each group are the same.\\
Secondly, \textbf{since there are a total of 8 sets of repeated paragraphs, the number of unique paragraphs is 8.} \\
Therefore, the answer to the question: 8. \\ \\

\colorbox[rgb]{0.8,0.8,0.8}{\textbf{Answer (OI):}} \\
\textbf{8.}
\end{tabular} \\ \\ \bottomrule
\end{tabular}
\end{table*}

\begin{table*}[t]
  \centering
  \footnotesize
\caption{\label{table15}
    Example outputs by FST for ASAP dataset. The \textbf{texts highlighted in bold} denote the key intermediate process generated by FST to solve this task.
  }
\begin{tabular}{l}
\toprule
\begin{tabular}[c]{@{}p{.96\textwidth}@{}}
\\
\colorbox[rgb]{0.8,0.8,0.8}{\textbf{Question:}} \\
Source Essay: ...\\
Describe the mood created by the author in the memoir. Support your answer with relevant and specific information from the memoir. \\
The above is the topic of the article. Now I will give you an article and the evaluation protocols. You need to score the article according to the evaluation protocols.\\ \\
The essay: ...\\ \\
The evaluation protocols:\\
Score 4: The response is a clear, complete, and accurate description of the mood created by the author. The response includes relevant and specific information from the memoir.\\
Score 3: The response is a mostly clear, complete, and accurate description of the mood created by the author. The response includes relevant but often general information from the memoir.\\
Score 2: The response is a partial description of the mood created by the author. The response includes limited information from the memoir and may include misinterpretations.\\
Score 1: The response is a minimal description of the mood created by the author. The response includes little or no information from the memoir and may include misinterpretations. Or the response relates minimally to the task.\\
Score 0: The response is incorrect or irrelevant or contains insufficient information to demonstrate comprehension.
\\ \\

\colorbox[rgb]{0.8,0.8,0.8}{\textbf{Answer (FT):}} \\
Simplification Task: \textbf{Regardless of the protocols, evaluate the essay and write a comment.} \\
Answer: The response mainly describes the mood created by the author in the memoir. Specifically, this response is given based on careful reading and understanding of the memoir. \textbf{It not only contains a description of the mood created by the author but also provides the reasons for deriving that description.} Therefore, it is a great response that meets the requirements of the task very well.\\ \\

\colorbox[rgb]{0.8,0.8,0.8}{\textbf{Answer (ST):}}\\
\textbf{According to the given comment, the response is of high quality because it gives a clear description of the mood created by the author in the memoir and the reasons for deriving that description.} Based on this comment and the specific evaluation protocols, the score of the response is 4. The response meets the requirement of score 4 in the evaluation protocols, which asks the response to contain a clear, complete, and accurate description of the mood created by the author, and relevant and specific information from the memoir.\\
Therefore, the score of the essay:\\
4. \\ \\

\colorbox[rgb]{0.8,0.8,0.8}{\textbf{Answer (OI):}} \\
A score of 4 is not the best rating result. The correct score of the essay is 3. The response gives a clear and right description of the mood created by the author in the memoir. However, A score of 4 also requires that the response includes relevant and specific information from the memoir. \textbf{Although the response includes an explanation of it, the explanation is not specific, such as citing content from the memoir and interpreting the author's experience.} \\
\textbf{Therefore, the score of the essay is 3.}
\end{tabular} \\ \\ \bottomrule
\end{tabular}
\end{table*}

\begin{table*}[t]
  \centering
  \footnotesize
\caption{\label{table16}
    Example outputs by FST for CommonGen-Hard dataset. The \underline{underlined texts} denote the words that must be included in the story, and the \textbf{texts highlighted in bold} denote the illogical content.
  }
\begin{tabular}{l}
\toprule
\begin{tabular}[c]{@{}p{.96\textwidth}@{}}
\\
\colorbox[rgb]{0.8,0.8,0.8}{\textbf{Question:}} \\
Write a short story containing all the following words:\\
vegetable, chess, keyboard, motorcycle, sun, sip, ski, part, shoe, mask, use, hose, racket, paint, wheel, hookah, hole, lawn, wand, base, calf, crash, game.
\\ \\

\colorbox[rgb]{0.8,0.8,0.8}{\textbf{Answer (FT):}} \\
Simplification Task: Generate a story belonging to a topic that is relevant to
the specified words. \\
Answer: \\
Topic: a garden party and some unexpected events occur. \\
Story: On a \underline{sunny} afternoon, the garden \underline{lawn} came alive with the sound of laughter $\cdots$ Friends was playing \underline{chess} $\cdots$ a \underline{hose} was being used to water the \underline{vegetable}. A few guests were relaxing with \underline{hookahs}, \underline{sipping} their favorite drinks. Others were playing basketball with \underline{rackets} $\cdots$ As the guests left, they each took a part of the day with them.
\\ \\

\colorbox[rgb]{0.8,0.8,0.8}{\textbf{Answer (ST):}}\\
To contain the given words, the revised story is: \\
As the \underline{sun} beat down on the lush \underline{lawn}, a group of friends gathered around an old, abandoned house rumored to be haunted. Their adventurous spirits were ignited as Mike arrived on his \underline{motorcycle}, ready to explore the mysterious abode $\cdots$ In one corner, hidden beneath an old \underline{shoe} and a \underline{mask}, was a mysterious \underline{hole} leading into the depths below $\cdots$ \textbf{They stumbled upon a \underline{keyboard} that seemed ancient} and a \underline{wand} emitting a faint glow. As they explored further, they encountered challenges. At one point, they found themselves playing a life-sized \underline{game} of \underline{chess} against a mysterious figure $\cdots$ Finally, they reached a breathtaking underground \underline{vegetable} garden. Exhausted yet exhilarated, they sat down to rest and \textbf{\underline{sip} from a \underline{hose}}. In a corner, they found \textbf{a \underline{hookah} gently releasing fragrant smoke into the air}. As they hung up their jackets and \underline{ski} equipment, they noticed a \underline{racket} and a \underline{paint} set $\cdots$\\ \\

\colorbox[rgb]{0.8,0.8,0.8}{\textbf{Answer (OI):}} \\
Step 1: The story misses some given words, such as part, use, racket, base, calf, and crash. \\
Step 2: The story contains some illogical contents, such as ``sip from a hose'', ``They stumbled upon a keyboard that seemed ancient'', and ``a hookah gently releasing fragrant smoke into the air''. \\
Therefore, the improved story is \\
This is a world full of magic. One day, as the \underline{sun} beat down on the lush \underline{lawn}, a group of friends gathered around an old, abandoned house, wanting to make it their \underline{base}. One of them, Mike arrived on his \underline{motorcycle}. Upon their arrival at the house, they were intrigued by a chessboard set up outside. They need to play \underline{chess}, moving the pieces into special formations to unlock the house. After success, they discovered a basement. In one corner, hidden beneath an old \underline{shoe} and a \underline{mask}, was a \underline{hole} leading into the depths below. They dove into the unknown. They stumbled upon a magic instrument \underline{keyboard} that seemed ancient, a \underline{wheel} etched with peculiar symbols, and a \underline{wand} emitting a faint glow. As they explored, they encountered challenges. They found themselves playing a life-sized \underline{game}, each move determining their fate. With teamwork, they emerged victorious. Finally, it seems they have entered a new world, one where magic doesn't exist. They found that although there was no magic, people \underline{used} tools instead, such as using a \underline{hose} to glue \underline{vegetables}, using a board to \underline{ski}, etc. They are trying new things, such as \underline{painting}, feeding a \underline{calf} grazing, hitting the ball with a \underline{racket}, \underline{sipping} a \underline{hookah}, etc. Suddenly, there was a \underline{crash}. A \underline{part} of the ground opened up. They are sucked into the hole and return to their world, and the previous experience is like a dream.
\end{tabular} \\ \\ \bottomrule
\end{tabular}
\end{table*}

\end{appendices}

\section*{Declarations}

\bmhead*{Funding} This research is supported by NSF of China (Nos: 62336003, 12371510), and NSF for Distinguished Young Scholar of Jiangsu Province (No: BK20220080).

\bmhead*{Conflict of interest} Yiliu Sun, Yanfang Zhang, Zicheng Zhao, Sheng Wan, and Chen Gong received research support from Nanjing University of Science and Technology (njust.edu.cn). Dacheng Tao received research support from The University of Sydney (ntu.edu.sg).

\bmhead*{Ethics approval and consent to participate} This research did not involve Human Participants and/or Animals. The authors do not foresee any ethical concerns with the content of the present paper.

\bmhead*{Consent for publication} The authors express their consent for publication.

\bmhead*{Data availability} All the datasets used in this research are available online. All their references are provided in this manuscript.

\bmhead*{Materials availability} All the materials used in this research are available online. All their references are provided in this manuscript.

\bmhead*{Code availability} The codes for the experiments conducted in this research can be found at the following URL: 
\href{https://github.com/sunyiliu/Fast-Slow-Thinking}{https://github.com/sunyiliu/Fast-Slow-Thinking}. 

\bmhead*{Author contribution} All authors contributed to the research conception and design, and commented on previous versions of the manuscript.

\bibliography{sn-bibliography}

\end{document}